\documentclass[lettersize,journal]{IEEEtran}
\usepackage{amsmath,amsfonts}
\usepackage{algorithmic}
\usepackage{algorithm}
\usepackage{array}
\usepackage[caption=false,font=normalsize,labelfont=sf,textfont=sf]{subfig}
\usepackage{textcomp}
\usepackage{stfloats}
\usepackage{url}
\usepackage{verbatim}
\usepackage{graphicx}
\usepackage{cite}
\usepackage{subfig}
\usepackage{xcolor}
\usepackage{booktabs}
\usepackage[citecolor=blue, colorlinks]{hyperref}
\usepackage{soul}
\usepackage{multirow}
\usepackage{makecell}
\usepackage{microtype}

\newcommand{\revise}[1]{\textcolor{black}{#1}}

\newcommand{\vstarup}{%
  \raisebox{-0.5ex}{
    \textsuperscript{%
      \kern-0.3pt%
      \setlength{\arraycolsep}{0pt}%
      \setlength{\extrarowheight}{-1pt}%
      $\begin{array}{c}* \\[-2.3pt] *\end{array}$%
    }%
  }%
}

\begin{document}



\title{Reflectance Prediction-based Knowledge Distillation for Robust 3D Object Detection in Compressed Point Clouds}

\author{Hao Jing, Anhong Wang, Yifan Zhang, Donghan Bu, and Junhui Hou, \textit{Senior Member, IEEE}

\thanks{This work was supported in part by the National Natural Science Foundation of China (U23A20314 and 62422118), in part by Industrial Application of Shanxi Provincial Technology Innovation Center (IVASXTIC2022), in part by Shanxi Province 'Reveal the List' Major Project (202301156401007), in part by Taiyuan Key Core Technology Research 'Reveal the List' Project (2024TYJB0128, 20240027), and in part by the Hong Kong Research Grants Council under Grants 11219324 and 11219422.  
}
\thanks{Hao Jing, Anhong Wang, and Donghan Bu are with the School of Electronic Information Engineering, Taiyuan University of Science and Technology, No. 66 Waliu Road, Taiyuan 030024, China. (e-mail: b20201591001@stu.tyust.edu.cn; ahwang@tyust.edu.cn; b202115310021@stu.tyust.edu.cn).}
\thanks{Yifan Zhang is with the School of Mechatronic Engineering and Automation, Shanghai University, Shanghai, China, and is also with the Department of Computer Science, City University of Hong Kong, Hong Kong, China (e-mail: yfzhang@shu.edu.cn).}
\thanks{Junhui Hou is with the Department of Computer Science, City University of Hong Kong, Hong Kong, China (e-mail: jh.hou@cityu.edu.hk).}
}

\markboth{Manuscript submitted to IEEE TIP}%
{Shell \MakeLowercase{\textit{et al.}}: A Sample Article Using IEEEtran.cls for IEEE Journals}


\maketitle

\begin{abstract}
\revise{Regarding intelligent transportation systems, low-bitrate transmission via lossy point cloud compression is vital for facilitating real-time collaborative perception among connected agents, such as vehicles and infrastructures, under restricted bandwidth.} In existing compression transmission systems, the sender lossily compresses point coordinates and reflectance to generate a transmission code stream, which faces transmission burdens from reflectance encoding and limited detection robustness due to information loss. To address these issues, this paper proposes a 3D object detection framework with reflectance prediction-based knowledge distillation (RPKD). We compress point coordinates while discarding reflectance during low-bitrate transmission, and feed the decoded non-reflectance compressed point clouds into a student detector. The discarded reflectance is then reconstructed by a geometry-based reflectance prediction (RP) module within the student detector for precise detection. A teacher detector with the same structure as the student detector is designed for performing reflectance knowledge distillation (RKD) and detection knowledge distillation (DKD) from raw to compressed point clouds. 
\revise{Our cross-source distillation training strategy (CDTS) equips the student detector with robustness to low-quality compressed data while preserving the accuracy benefits of raw data through transferred distillation knowledge.}
\revise{Experimental results on the KITTI and DAIR-V2X-V datasets} demonstrate that our method can boost detection accuracy for compressed point clouds across multiple code rates. 
We will release the code publicly at \url{https://github.com/HaoJing-SX/RPKD}.
\end{abstract}


\section{Introduction}
\IEEEPARstart{L}{iDAR} point clouds, composed of sparse points with both geometric (coordinates) and attribute (reflectance) information about object surfaces, are vital for predicting the categories, positions, and bounding boxes of 3D objects. \revise{In vehicle-to-everything (V2X) cooperative perception, receivers need to rapidly acquire LiDAR point clouds captured by neighboring senders to enable an expanded perception range and accurate object detection \cite{chen2019cooper,ye2020cooperative,xu2023v2v4real,bai2024survey}.} Raw point clouds are extremely voluminous and require a high transmission bandwidth, making it difficult to realize real-time cooperative perception. To this end, point cloud compression reduces the amount of transmitted data, improving data acquisition efficiency at the receiver. However, compressed point clouds generated by existing methods often lose some geometric coordinates and reflectance information, leading to suboptimal performance in 3D object detection. Therefore, enhancing the 3D detector’s robustness for compressed point clouds is of great research significance.

\begin{figure*}[!htb]
\begin{center}
	\centering
	\subfloat[]{
		\includegraphics[width=3.6in]{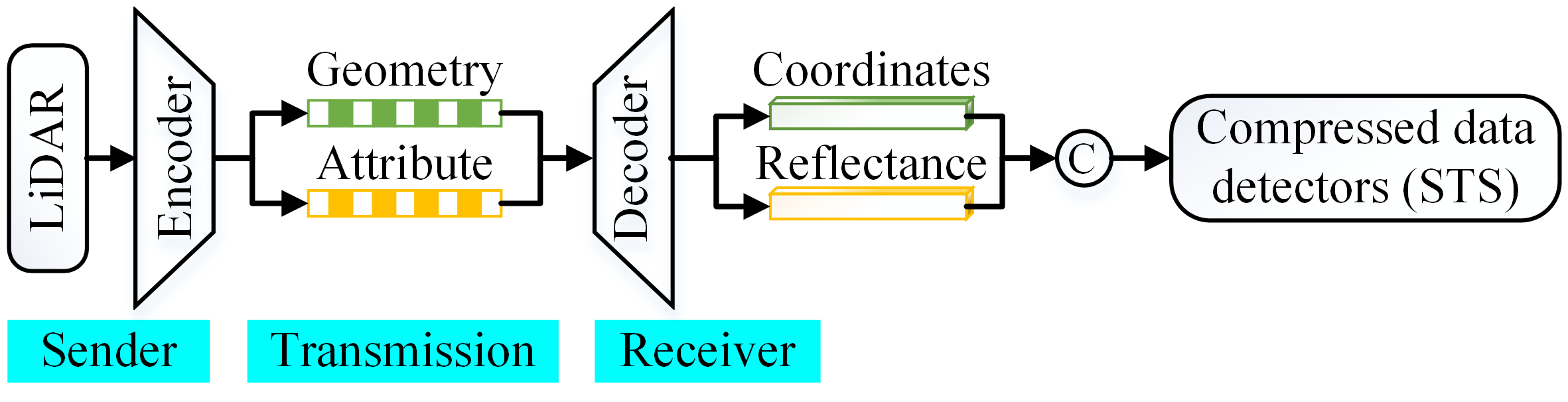}
		\label{fig:1a}}
	\hfil
	\subfloat[]{
		\includegraphics[width=3.0in]{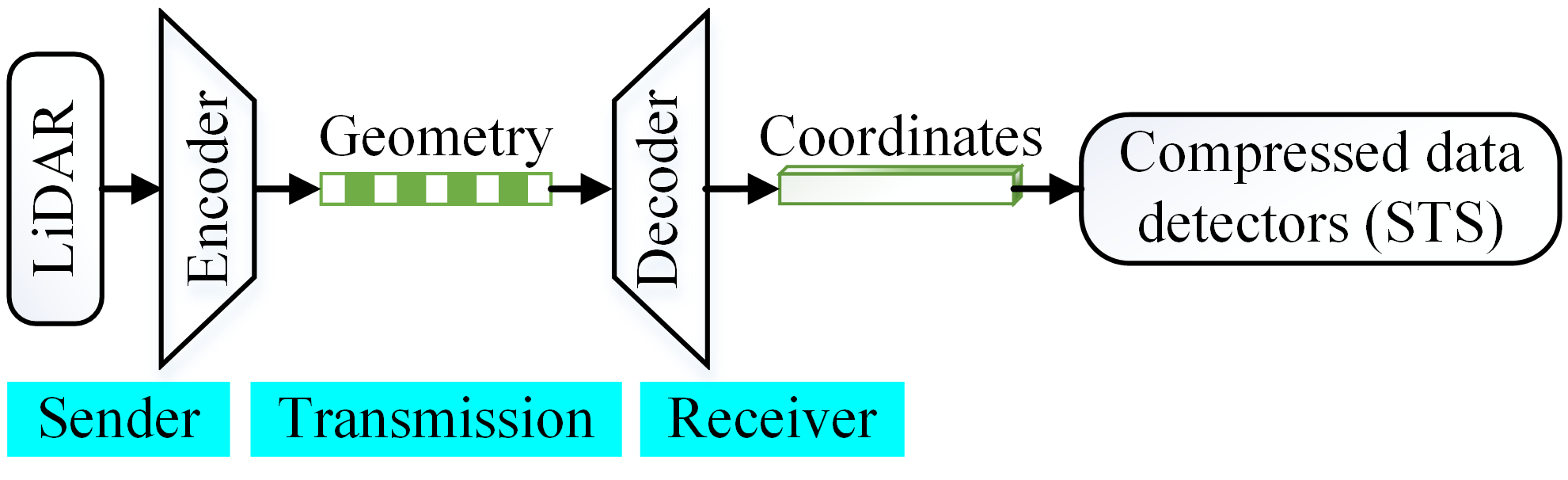}
		\label{fig:1b}}	
	\vspace{-0.1cm} 
	\subfloat[]{
		\includegraphics[width=2.8in]{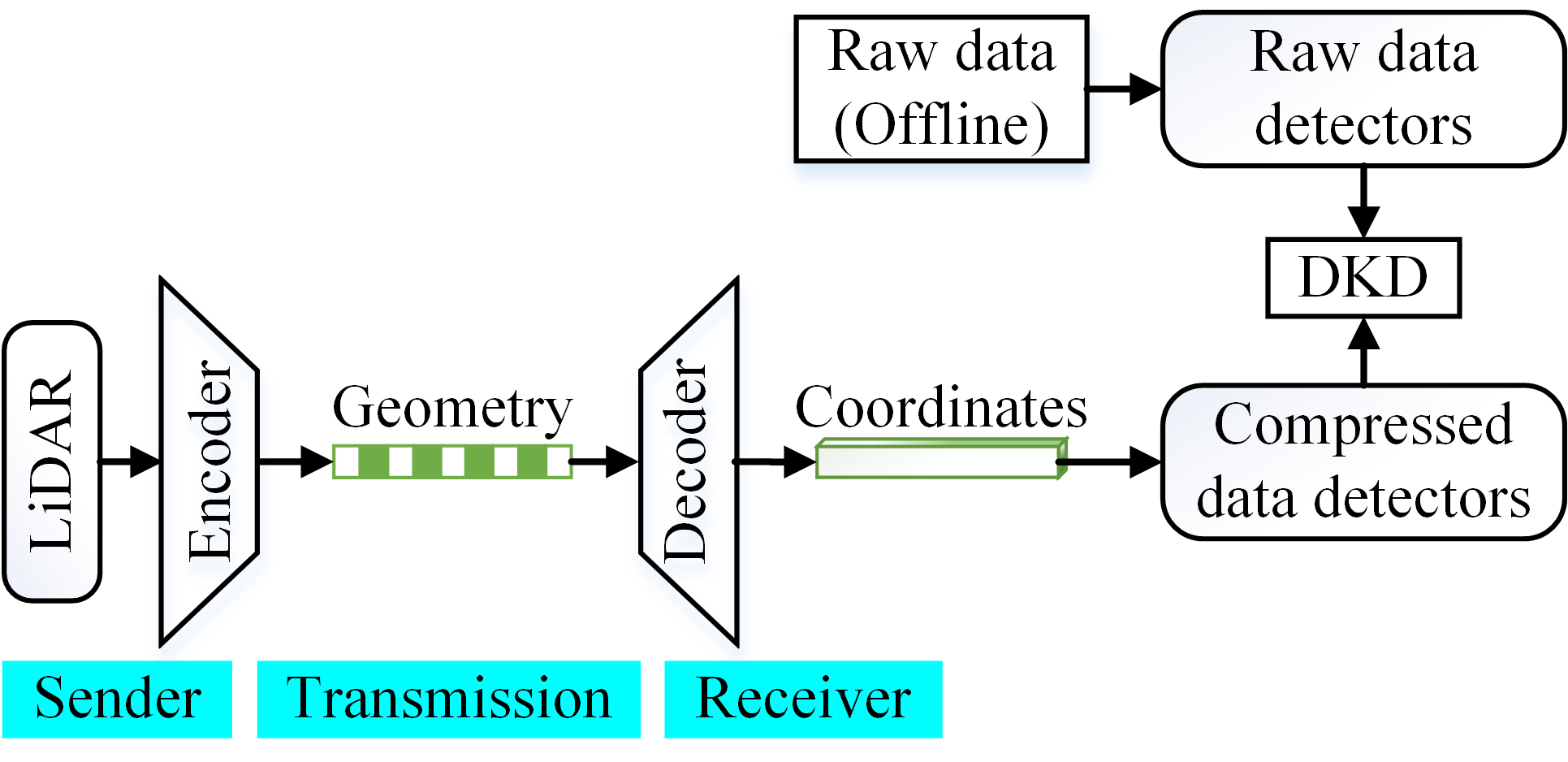}
		\label{fig:1c}}
	\hfil
	\subfloat[]{
		\includegraphics[width=3.8in]{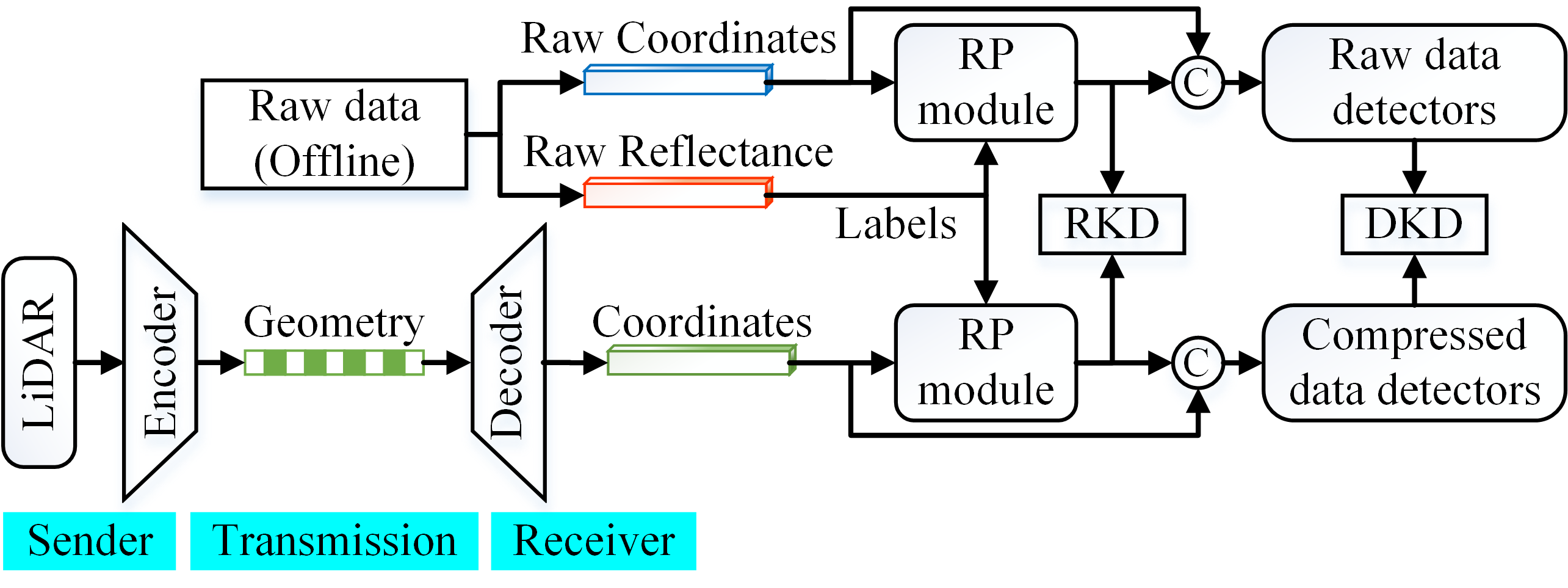}
		\label{fig:1d}}
	
	\caption{\revise{Comparison of 3D object detection methods for compressed LiDAR point clouds. (a) With reflectance encoding, detection suffers from the transmission burden of the attribute bitstream. (b) Without reflectance encoding, only geometric coordinates are compressed, improving efficiency but showing poor robustness under a single-source training strategy (STS). (c) Building on raw and non-reflectance compressed data, SparseKD (CDTS) transfers detection knowledge from raw to compressed data detectors, yielding slight performance gains. (d) Our RPKD (CDTS) introduces the RP module and RKD constraint to recover reflectance, significantly boosting detection performance on non-reflectance compressed data.}}
	\label{fig:figure1}
\end{center}
\end{figure*}

Existing 3D object detection methods in compressed point clouds can be classified into two categories based on their reflectance processing approaches: \revise{3D object detection with \cite{cao2021compression, wang2024versatile,wang2024versatile2} or without reflectance encoding \cite{que2021voxelcontext,chen2022point,fu2022octattention,lodhi2023sparse,wang2022sparse,you2025reno}.} For 3D object detection with reflectance encoding, both G-PCC \cite{cao2021compression} and Unicorn \cite{wang2024versatile,wang2024versatile2} compressed voxelized point-cloud coordinates together with reflectance, generating geometry and attribute bitstreams for reconstruction and detection tasks, as shown in Fig. \hyperref[fig:figure1]{\ref{fig:figure1}(a)}. \revise{However, the attribute bitstream in these methods introduces a transmission burden that multiplies with the number of connected vehicles and infrastructures.}
To solve this issue, in 3D object detection without reflectance encoding, existing compression methods based on voxelized point clouds and deep learning \cite{que2021voxelcontext,chen2022point,fu2022octattention,lodhi2023sparse,wang2022sparse,you2025reno} typically focused on compressing geometric coordinates while discarding reflectance to elevate transmission efficiency, as presented in Fig. \hyperref[fig:figure1]{\ref{fig:figure1}(b)}. 
However, lossy compression by these methods causes substantial information loss in low-bitrate transmission, severely impacting 3D object detection accuracy. \revise{To address these challenges, this paper adopts the voxel-based lossy geometric compression method PCC-S \cite{chen2022point} to maximize bandwidth reduction and systematically investigates training strategies for non-reflectance compressed point clouds to develop a robust detector.}

\revise{The existing raw-data single-source training strategy (STS-R) \cite{que2021voxelcontext,wang2022point} pretrained detection models on raw data and evaluated them on low-quality compressed data, resulting in limited robustness. To handle the low quality of compressed point clouds, we adopt a compressed-data single-source training strategy (STS-C), training and evaluating at each compression rate to improve detection performance. In 3D object detection knowledge distillation, FD \cite{shan2023focal} transferred classification, regression, and feature knowledge from high-resolution to low-resolution data, thereby enhancing detection accuracy on low-resolution inputs. Building on this idea, we propose a cross-source distillation training strategy (CDTS), in which raw and compressed point-cloud models form a teacher-student model pair, and response-based detection knowledge distillation (DKD) from SparseKD \cite{yang2022towards} is applied to transfer accurate detection knowledge, as illustrated in Fig. \hyperref[fig:figure1]{\ref{fig:figure1}(c)}.}

\begin{figure}[!htb]
\begin{center}
    \begin{tabular}{c@{\hspace{-4mm}}}
    \hspace{-15pt}\includegraphics[width=3.5in]{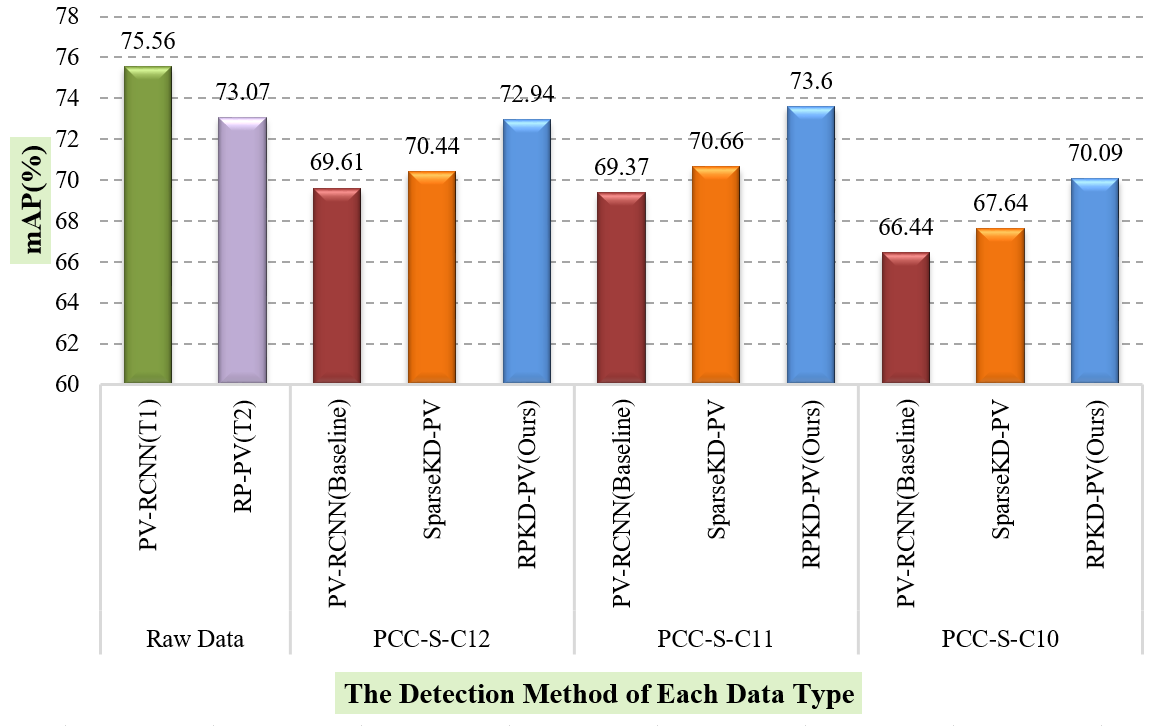}   \\
\end{tabular}
  \caption{\revise{Quantitative comparison of different detection methods on multiple KITTI compressed data. STS-R and STS-C denote single-source training strategies based on raw and compressed data, respectively, while CDTS refers to the cross-source distillation training strategy proposed in this paper. The blue dotted line shows the detection performance of the PV-RCNN baseline on raw data.}}
  \label{fig:figure2}
\end{center}
\end{figure}

\revise{Using PV-RCNN as the backbone, Fig. \ref{fig:figure2} reports performance for different training strategies and detection methods on non-reflectance compressed data. SparseKD-PV (CDTS) outperforms PV-RCNN (STS-C) by an average of 1.11 mAP across various data types, yet still lags behind the PV-RCNN baseline on raw data. In general, LiDAR reflectance offers critical cues for 3D object recognition, thereby enhancing detection reliability. For SparseKD-PV (CDTS), however, discarding reflectance to improve transmission efficiency remains a key factor contributing to the suboptimal performance. To overcome this limitation, we introduce a dedicated reflectance prediction (RP) module at the receiver and employ knowledge distillation techniques to refine reflectance estimation, ultimately boosting detection performance on compressed point clouds.}

Incorporating the previous analysis, we propose a 3D object detection framework with reflectance prediction-based knowledge distillation (RPKD) to improve detection accuracy for non-reflectance compressed point clouds, as depicted in Fig. \hyperref[fig:figure1]{\ref{fig:figure1}(d)}. Since compressed and raw point clouds lack direct point-to-point correspondence, we design a reflectance cross-match (RCM) module to assign reflectance labels from raw point-cloud voxels to compressed points. To construct suitable knowledge for reflectance distillation, we introduce a comparable reflectance inter-match (RIM) module to generate reflectance labels for the raw point-cloud detector. Besides, the proposed reflectance prediction (RP) module comprehensively extracts geometric features of voxelized point clouds, achieving reflectance reconstruction at the receiver. \revise{Our cross-source distillation training strategy (CDTS) utilizes response-based reflectance knowledge distillation (RKD) and detection knowledge distillation (DKD) to transfer guided knowledge from raw to compressed point-cloud detectors, enabling more accurate reflectance reconstruction and improved 3D object detection. As shown in Fig. \ref{fig:figure2}, RPKD-PV (CDTS) significantly raises the mAP values across various data types, demonstrating the effectiveness of our method for non-reflectance compressed point clouds.}

In summary, the contributions of this work are as follows.
\begin{itemize}
  \item We propose an RCM module to generate reflectance labels for compressed points based on the corresponding raw point-cloud voxels. 
  \item  At the receiver, we design a geometry-based RP module to reconstruct discarded reflectance, thereby improving detection performance for non-reflectance compressed point clouds.
  \item  \revise{The proposed CDTS transfers supervisory knowledge from raw to compressed point-cloud detectors using two response-based distillation constraints, RKD and DKD, further enhancing detection robustness for low-quality compressed data.}
\end{itemize}

The remainder of the paper is organized as follows. Section~\ref{sec:related_work} reviews existing studies in terms of 3D object detection, point cloud compression, and knowledge distillation, highlighting relevant technologies. In Section~\ref{sec:approach}, we present the overall architecture of a 3D object detection framework with reflectance prediction-based knowledge distillation and elaborate on its key components. \revise{Section~\ref{sec:experiments} demonstrates the effectiveness of our method on the KITTI and DAIR-V2X-V datasets, and includes ablation studies to assess the impact of each component.} Finally, Section~\ref{sec:conclusion} concludes this paper.

\section{Related Work}
\label{sec:related_work} 
\subsection{3D Object Detection}
In 3D object detection for raw point clouds, existing methods can be sorted into two classes according to the data format of network input: point-based \cite{shi2019pointrcnn,chen2022sasa,zhang2022not,zhang2023unleash,he2023sa,feng2020relation} 
and voxel-based \cite{shi2020pv,deng2021voxel,yan2018second,shi2023pv,lang2019pointpillars,yin2021center,wu2023transformation,zhang2023glenet,lee2024re,xiao2023balanced,9870669,xie2023x} methods. Point-based detectors extract hierarchical point-wise features by various sampling and aggregation methods to predict 3D objects. In contrast, voxel-based detectors utilize 3D sparse convolution to extract voxel features, and these features are projected onto the bird's eye view (BEV) for 3D object detection. In 3D object detection for compressed point clouds, those generated by range image-based \cite{feng2020real,wang2022point,wang2022r,zimmer2024pointcompress3dpointcloudcompression} and voxel-based \cite{cao2021compression,wang2024versatile,wang2024versatile2,que2021voxelcontext,chen2022point,fu2022octattention,lodhi2023sparse,wang2022sparse,you2025reno} compression frameworks were input into detectors for recognition and localization. These methods typically rely on raw point clouds for pretraining 3D detectors and evaluate detection performance on lossy compressed point clouds at various code rates using the pretrained models. \revise{However, the raw-data single-source training strategy lacks adaptability to low-quality compressed data, resulting in a significant decline in detection accuracy. To address this, we propose a novel CDTS that improves detector robustness to low-quality compressed data while inheriting the accuracy of raw data.}

\subsection{Point Cloud Compression}
LiDAR point cloud compression methods can be categorized into three classes based on data representation: point-based \cite{he2022density,cai2024voloc,li2024point}, image-based \cite{feng2020real,wang2022point,wang2022r,zhao2022real,zhao2022real2,zimmer2024pointcompress3dpointcloudcompression,zhang2023flattening,zeng2024dynamic,zhang2022reggeonet}, and voxel-based \cite{cao2021compression,wang2024versatile,wang2024versatile2,que2021voxelcontext,chen2022point,fu2022octattention,lodhi2023sparse,wang2022sparse,you2025reno,8816692} methods. Point-based methods utilize flexible point sampling and feature extraction to generate point-wise features, achieving point cloud compression and reconstruction. However, these methods are confronted with uneven 3D object loss and noise-point interference due to the lack of spatial structure and context information during lossy compression. Image-based methods project point clouds onto 2D images for quantization and encoding, but they introduce more severe three-dimensional spatial distortion and reconstruction errors compared to voxel-based methods. By analyzing octree or multi-scale sparse tensors, voxel-based methods enhance compression performance via context feature learning, multi-scale joint optimization, and redundant encoding reduction. Among these compression methods, only G-PCC \cite{cao2021compression} and Unicorn \cite{wang2024versatile2} incorporated reflectance encoding modules, as reflectance encoding presents challenges in structural design, end-to-end optimization, and data-flow overhead. To solve this problem, we design an RP module at the receiver to supplement reflectance information for detection training on non-reflectance compressed point clouds generated by voxel-based methods. Our approach effectively alleviates the design complexity of LiDAR point cloud compression and provides an end-to-end optimization strategy for reflectance prediction at the receiver.

\begin{figure*}[b]
\begin{center}
    \begin{tabular}{c@{\hspace{-4mm}}}
  \includegraphics[width=0.95\textwidth]{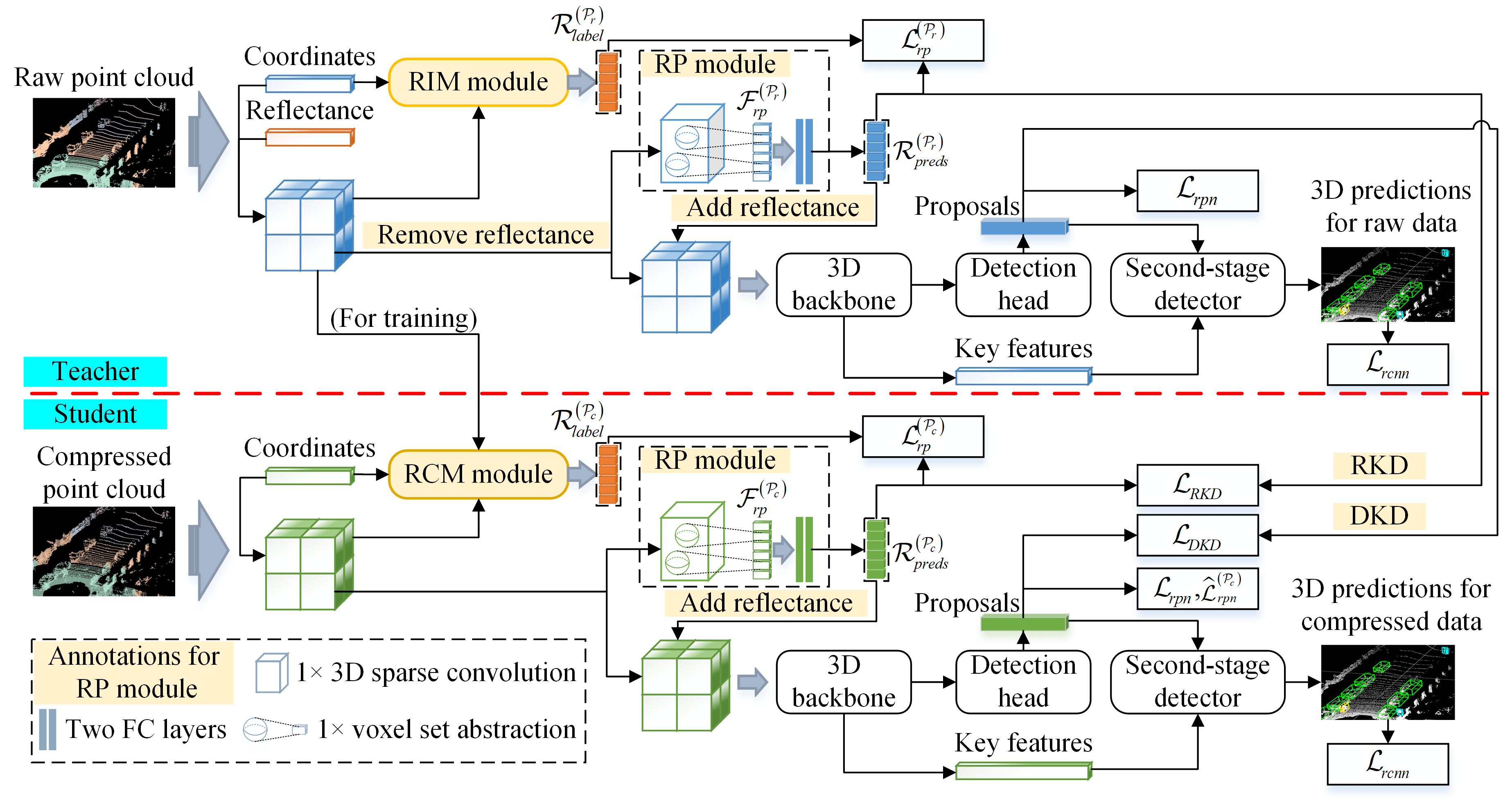}   \\
\end{tabular}
  \caption{Overview of our method. During student detector training, the RCM module assigns reflectance labels from raw point-cloud voxels to compressed points, while the RP module extracts geometric features to predict compressed point reflectance. In contrast, the teacher detector’s RP module predicts raw point reflectance based on labels obtained from the RIM module. The predicted reflectance is then integrated into non-reflectance voxels for detection learning. \revise{Building on CDTS, RKD transfers guided reflectance knowledge to the student detector, while DKD utilizes first-stage proposals to convey accurate detection knowledge.}}
  \label{fig:figure3}
\vskip -10pt
\end{center}
\end{figure*}

\subsection{Knowledge Distillation}
Knowledge distillation methods for 3D object detection can be divided into three categories according to their research objectives: model compression distillation \cite{yang2022towards,zhang2023pointdistiller,li2023representation,zhang2024instkd,cho2023itkd,chen2024graph}, detection improvement distillation \cite{ju2022paint,gambashidze2024weak}, and weak-data robustness distillation \cite{wei2022lidar,shan2023focal,huang2024sunshine}. 
In model compression distillation, SparseKD \cite{yang2022towards} introduced response-based constraints through Logit and Label KD to guide lightweight student models, achieving accurate and efficient 3D object detection. In detection improvement distillation, SPNet~\cite{ju2022paint} assigned category values to the interior points of ground truth (GT) boxes, while X-Ray~\cite{gambashidze2024weak} addressed object point loss caused by sparsity and occlusion utilizing tracking information from consecutive-frame point clouds. Both methods augment data representations with strong semantic priors, providing valuable supervisory knowledge for distillation. In weak-data robustness distillation, student models employed low-quality point clouds from rainy conditions \cite{huang2024sunshine} or low-resolution sensor data \cite{wei2022lidar,shan2023focal}, learning complete and accurate knowledge from raw point clouds to improve detection robustness in weak-data scenarios. Inspired by weak-data robustness distillation methods, our RPKD framework treats non-reflectance compressed point clouds in vehicle networking data transmission as weak data for student models, with raw point clouds guiding teacher models, aiming to develop robust compressed point-cloud detectors.

\subsection{V2X Collaborative Perception}
\revise{To overcome single-vehicle perception limitations, V2X collaborative perception \cite{xu2023v2v4real,bai2024survey,yang2024v2x,wang2025collaborative,yu2022dair} develops early, intermediate, and late fusion strategies based on shared information among connected agents. These strategies correspond to the transmission of raw data, intermediate layer features, and prediction results, respectively. In early fusion with LiDAR \cite{chen2019cooper,ye2020cooperative}, raw point clouds offer precise scene information and are easily registered, but they require significant transmission bandwidth. Cooper \cite{chen2019cooper} mitigated the bandwidth demand for mutual transmission between two vehicles by extracting key point clouds using RoIs. However, this approach still faces challenges in scaling to large V2X collaborative scenarios. To address these issues, this paper pioneeringly introduces voxel-based lossy geometric compression into collaborative perception, effectively balancing low bandwidth and high accuracy. Regarding datasets, DAIR-V2X \cite{yu2022dair} is more mature and was released earlier than V2X-Radar \cite{yang2024v2x}. For detection tasks, we focus on single-view 3D detection in DAIR-V2X-V and evaluate performance on compressed data across various object categories, thus laying the foundation for future collaborative detection.}

\section{Proposed Approach}
\label{sec:approach}
\subsection{Overview of our Framework}
\revise{In collaborative perception with LiDAR point cloud compression, reflectance encoding imposes a linearly increasing transmission burden as the number of connected agents grows.}
According to the separable compression mechanism of G-PCC~\cite{cao2021compression}, existing methods typically disregard reflectance compression and focus solely on optimizing geometric coordinate encoding to achieve efficient transmission and accurate detection. As a result, decoded compressed point clouds often lack reflectance information and suffer from point number reduction and reconstruction errors, which limit 3D object detection performance. To address this issue, we propose a 3D object detection framework with reflectance prediction-based knowledge distillation (RPKD), as shown in Fig. \ref{fig:figure3}, to obtain more robust and precise 3D detectors against non-reflectance compressed point clouds.

First, our reflectance cross-match (RCM) module correlates lossy compressed points with neighboring raw point-cloud voxels. Average reflectance of the nearest voxel is assigned as the reflectance label for the current compressed point. Similarly, the reflectance inter-match (RIM) module assigns the average reflectance of a voxel as the label for corresponding raw points. Next, the proposed reflectance prediction (RP) module generates point-wise geometric features using 3D sparse convolution and voxel set extraction to predict reflectance. The predicted reflectance is then integrated into the corresponding non-reflectance voxels for subsequent detection tasks. \revise{Finally, our cross-source distillation training strategy (CDTS) transfers guiding knowledge from raw to compressed point-cloud detectors through two response-based distillation constraints.} Specifically, reflectance knowledge distillation (RKD) and detection knowledge distillation (DKD) concentrate on the RP module’s reflectance knowledge and the first-stage proposal knowledge, respectively.
In what follows, we detail the proposed framework.

\begin{figure}[!htb]
\begin{center}
    \begin{tabular}{c@{\hspace{-4mm}}}
    \hspace{-15pt}\includegraphics[width=3in]{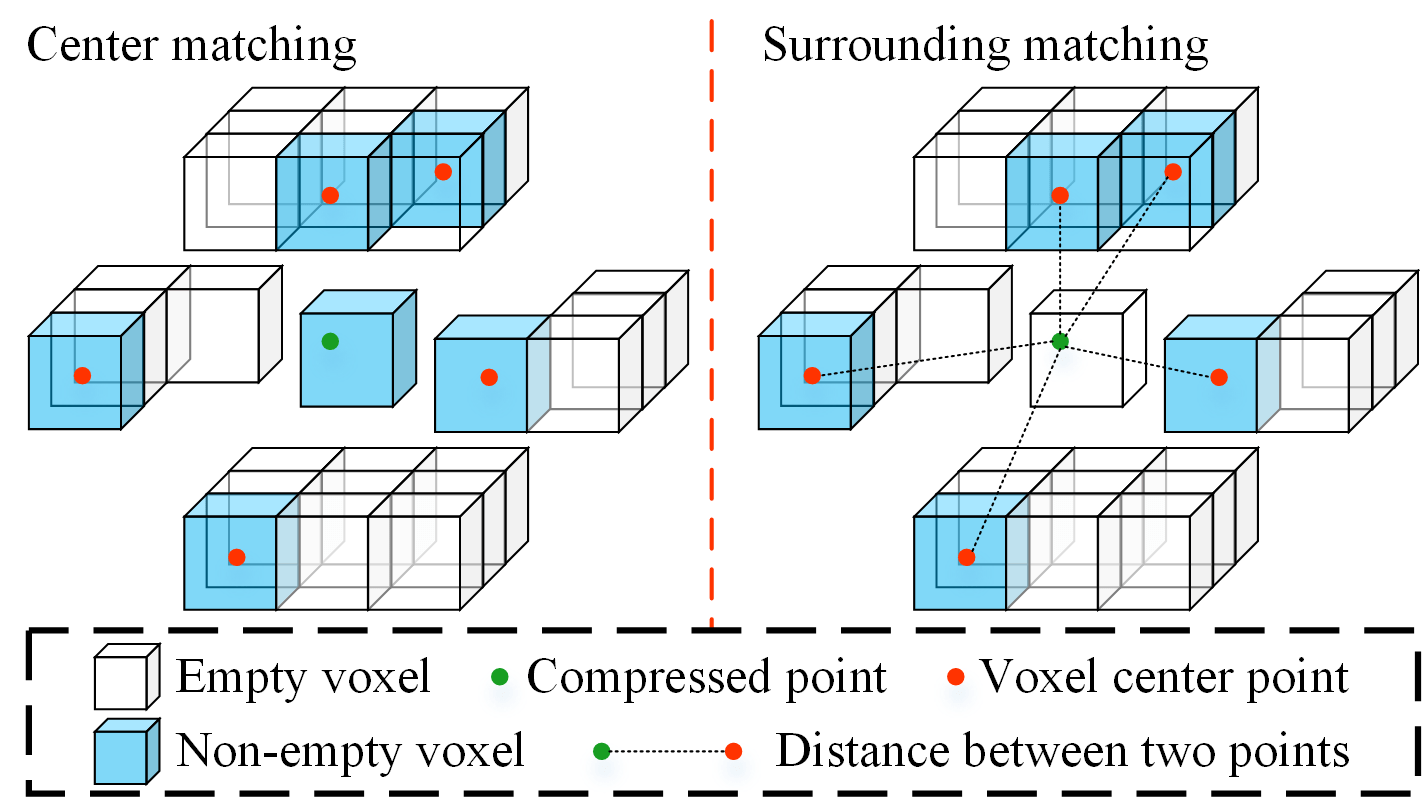}   \\
\end{tabular}
  \caption{Two matching methods of the RCM module. All voxels originate from voxelizing raw point clouds. If a compressed point lies within a non-empty voxel, the voxel’s mean reflectance is assigned to it. Otherwise, we search for the surrounding non-empty voxels within a 1-voxel matching range and use the nearest voxel’s mean reflectance as the matching value for this compressed point.}
  \label{fig:figure4}
\vskip -10pt
\end{center}
\end{figure}

\subsection{Reflectance Label Generation}\label{section:3B}
Due to point number reduction and reconstruction errors from voxel-based lossy compression, compressed points do not correspond one-to-one with raw points. For compressed point reflectance prediction, we introduce an RCM module based on spatial relationships to assign reflectance labels from raw point-cloud voxels to corresponding compressed points. These labels, used as prior knowledge for reflectance prediction, are generated frame by frame during detection training and do not need to be compressed or transmitted. Additionally, to transfer effective knowledge during reflectance distillation, we also design a similar RIM module to acquire reflectance labels for raw points. The principles of these modules are outlined below. 

\textbf{Reflectance Cross-match Module.} For geometric coordinate encoding, voxel-based compression methods \cite{wang2024versatile,chen2022point} are often employed to obtain non-reflectance compressed point clouds. With each compressed point’s reflectance assigned to zero, raw and compressed point clouds fed into our detection networks are represented as ${{\cal P}_r}$ and ${{\cal P}_c}$. These point clouds are voxelized to the same scale, generating raw point-cloud voxels ${{\cal V}^{\left( {{{\cal P}_r}} \right)}}$ and compressed point-cloud voxels ${{\cal V}^{\left( {{{\cal P}_c}} \right)}}$. When using the compressed point coordinates 
and the voxels of raw and compressed point clouds as inputs, the principle of our RCM module can be described as follows.


As depicted in Fig. \ref{fig:figure4}, there are two spatial matching strategies between compressed points and raw point-cloud voxels: center matching and surrounding matching. Specifically, the center voxel corresponding to compressed point coordinates ${c_{ci}}$ is denoted as $v_{center}^{\left( {{c_{ci}}} \right)}$, and the surrounding non-empty voxels within a 1-voxel matching range are ${\cal V}_s^{\left( {{c_{ci}}} \right)} \in {{\cal V}^{\left( {{{\cal P}_r}} \right)}}$. The initial reflectance labels for compressed points $\widehat {\cal R}_{label}^{\left( {{{\cal P}_c}} \right)}$ are defined as:
\begin{equation}
\label{1}
\widehat {\cal R}_{label}^{\left( {{{\cal P}_c}} \right)} = \begin{cases}
r_{m}\left( v_{center}^{\left( c_{ci} \right)} \right), & v_{center}^{\left( c_{ci} \right)} \in {\cal V}^{\left( {\cal P}_r \right)}, \\
r_{m}\left( \arg \min d\left( c_{ci}, {\cal V}_s^{\left( c_{ci} \right)} \right) \right), &v_{center}^{\left( c_{ci} \right)} \notin {\cal V}^{\left( {\cal P}_r \right)},
\end{cases}
\end{equation}
where ${r_m}\left( {v_{center}^{\left( {{c_{ci}}} \right)}} \right)$ represents the average reflectance of a center voxel,
and ${r_m}\left( {\arg \min d\left( {{c_{ci}},{\cal V}_s^{\left( {{c_{ci}}} \right)}} \right)} \right)$ is the nearest voxel’s average reflectance in surrounding matching. However, when multiple compressed points exist within a compressed point-cloud voxel, their reflectance labels may come from different raw point-cloud voxels, resulting in significant numerical discrepancies that negatively impact compressed point reflectance prediction. To address this issue, we further calculate the average reflectance labels for all points within each compressed point-cloud voxel. The final reflectance labels for compressed points ${\cal R}_{label}^{\left( {{{\cal P}_c}} \right)}$ are thus expressed as:
\begin{equation}
\label{2}
\begin{array}{*{20}{l}}
{{\cal R}_{label}^{\left( {{{\cal P}_c}} \right)} = \left[ {\left. {{{\cal M}_c}\left( {\widehat r_{label}^{\left( {{c_{ci}}} \right)},{c_{ci}},{v_{cj}}} \right)} \right|\widehat r_{label}^{\left( {{c_{ci}}} \right)} \in \widehat {\cal R}_{label}^{\left( {{{\cal P}_c}} \right)}} \right],}\\
{i \in \{ 1, \cdots ,{n_c}\} ,j \in \{ 1, \cdots ,{n_{vc}}\} ,}
\end{array}
\end{equation}
where $\widehat r_{label}^{\left( {{c_{ci}}} \right)}$ denotes the reflectance label for each point within the current compressed point-cloud voxel ${v_{cj}}$, and ${{\cal M}_c}\left(  \cdot  \right)$ represents the averaging operation applied to these reflectance labels, with $n_c$ and $n_{vc}$ denoted as the corresponding numbers of compressed points and voxels, respectively.

\textbf{Reflectance Inner-match Module.} In the RPKD framework, it is crucial to establish suitable reflectance knowledge from raw point clouds to guide the prediction process for compressed point-cloud reflectance. Specifically, we treat the reflectance predictions for raw and compressed point clouds as a teacher-student pair, where both sets of labels need to maintain spatial correspondence and information similarity. In the RCM module, the reflectance labels for compressed points are derived from the corresponding raw point-cloud voxels. Therefore, our RIM module avoids directly using raw point reflectance and instead assigns the average reflectance of raw point-cloud voxels to corresponding points. Since raw points correspond exactly to their voxels, the RIM module uses center matching to obtain the raw point reflectance labels ${\cal R}_{label}^{\left( {{{\cal P}_r}} \right)}$.


\revise{These pluggable matching modules are designed for non-reflectance compressed point clouds, which are obtained through voxel-based geometric compression methods such as PCC-S. When compressed point clouds inherently include reflectance, the reflectance labels are directly available.}

\subsection{Reflectance Prediction Module}
Concerning the reflectance reconstruction of compressed point clouds, Unicorn \cite{wang2024versatile2} utilized current-scale reflectance residual features and low-scale reflectance priors through a DNN module to predict reflectance. In contrast, our RP module exploits the geometric features of non-reflectance compressed point clouds and the reflectance priors of raw point clouds to reconstruct reflectance at the receiver. \revise{In collaborative perception with LiDAR point cloud compression, the concept of reflectance prediction is crucial for ensuring detection robustness against low-quality compressed data, while eliminating reflectance encoding to push the boundaries of transmission bandwidth reduction.}
As shown in Fig. \ref{fig:figure3}, the structure and principle of our RP module are detailed as follows.

\textbf{RP Module for Compressed Point Clouds.} In the compressed point-cloud detector, the RP module takes non-reflectance compressed point-cloud voxels
as input. To accurately predict reflectance for compressed points, the RP module employs 3D sparse convolution to extract voxel features and voxel set extraction to refine point-wise features. We define the geometric features of compressed points ${\cal F}_{rp}^{\left( {{{\cal P}_c}} \right)}$ as:
\begin{equation}
\label{4}
\begin{array}{*{20}{l}}
{{\cal F}_{rp}^{\left( {{{\cal P}_c}} \right)} = {\cal V}{\cal S}{{\cal A}^{\left( {1 \times } \right)}}\left( {{c_{ci}}{\rm{,SCon}}{{\rm{v}}^{\left( {1 \times } \right)}}\left( {{\cal T}\left( {{{\cal V}^{\left( {{{\cal P}_c}} \right)}}} \right)} \right)} \right),}\\
{i \in \{ 1, \cdots ,{n_c}\} ,}
\end{array}
\end{equation}
where ${\cal T}\left(  \cdot  \right)$ denotes a sparse tensor transformation module, ${\rm{SConv}}{\left(  \cdot  \right)^{\left( {1 \times } \right)}}$ is the 3D sparse convolution with a ${1 \times}$ downsampling size (i.e., the initial size) in the voxel-wise basic backbone, and ${\cal V}{\cal S}{\cal A}{\left(  \cdot  \right)^{\left( {1 \times } \right)}}$ represents the ${1 \times}$ voxel set extraction corresponding to ${\rm{SConv}}{\left(  \cdot  \right)^{\left( {1 \times } \right)}}$. The voxel set extraction structure in our RP module is identical to that in PV-RCNN \cite{shi2020pv}, but the number of sampling points is equal to the number of compressed points in the current frame. Since the reflectance input to the detection network is a single-channel floating-point scalar in the range of $\left[ {0,1} \right)$, we use two fully connected (FC) layers as a prediction head to generate compressed point reflectance.
The predicted reflectance is then incorporated into the corresponding compressed point-cloud voxels based on their spatial relationship for subsequent training and detection.

We screen the compressed point reflectance labels with values greater than zero, and define their indices as ${I_c} = \left\{ {\left. i \right|r_{label}^{\left( {{c_{ci}}} \right)} > 0,r_{label}^{\left( {{c_{ci}}} \right)} \in {\cal R}_{label}^{\left( {{{\cal P}_c}} \right)} } \right\}$.
Mean Squared Error Loss (MSELoss) is applied to constrain compressed point reflectance prediction, and the corresponding loss function ${\cal L}_{rp}^{\left( {{{\cal P}_c}} \right)}$ is expressed as follows:
\begin{equation}
\label{5}
{\cal L}_{rp}^{\left( {{{\cal P}_c}} \right)} = \frac{1}{{\left| {{I_c}} \right|}}\sum\limits_{i \in {I_c}} {{{\left( {r_{preds}^{\left( {{c_{ci}}} \right)} - r_{label}^{\left( {{c_{ci}}} \right)}} \right)}^2}} ,i \in \{ 1, \cdots ,{n_c}\} ,
\end{equation}
where $\left| {{I_c}} \right|$ represents the number of compressed point reflectance labels with values greater than zero, and $r_{preds}^{\left( {{c_{ci}}} \right)}$ denotes the predicted reflectance for compressed points.

\textbf{RP Module for Raw Point Clouds.} In 3D object detection knowledge distillation, maintaining similar network structures between teacher and student detectors benefits the efficient transfer of teacher knowledge. Therefore, in this study, we design an analogous RP module in the raw point-cloud detector to predict raw point reflectance. Non-reflectance raw point-cloud voxels 
are fed into this module, and the geometric features of raw points ${\cal F}_{rp}^{\left( {{{\cal P}_r}} \right)}$ are represented as:
\begin{equation}
\label{6}
\begin{array}{*{20}{l}}
{{\cal F}_{rp}^{\left( {{{\cal P}_r}} \right)} = {\cal V}{\cal S}{{\cal A}^{\left( {1 \times } \right)}}\left( {{c_{rk}}{\rm{,SCon}}{{\rm{v}}^{\left( {1 \times } \right)}}\left( {{\cal T}\left( {{{\cal V}^{\left( {{{\cal P}_r}} \right)}}} \right)} \right)} \right),}\\
{k \in \{ 1, \cdots ,{n_r}\} .}
\end{array}
\end{equation}
where $c_{rk}$ represents raw point coordinates, and $n_r$ is the number of raw points. Subsequently, we predict raw point reflectance and assign it to corresponding voxels, resulting in raw point-cloud voxels with recovered reflectance.

We screen the raw point reflectance labels that are greater than zero and define their indices as 
${I_r} = \left\{ {\left. k \right|r_{label}^{\left( {{c_{rk}}} \right)} > 0,r_{label}^{\left( {{c_{rk}}} \right)} \in {\cal R}_{label}^{\left( {{{\cal P}_r}} \right)}} \right\}$.
The loss function for raw point reflectance prediction ${\cal L}_{rp}^{\left( {{{\cal P}_r}} \right)}$ is defined as:
\begin{equation}
\label{7}
{\cal L}_{rp}^{\left( {{{\cal P}_r}} \right)} = \frac{1}{{\left| {{I_r}} \right|}}\sum\limits_{k \in {I_r}} {{{\left( {r_{preds}^{\left( {{c_{rk}}} \right)} - r_{label}^{\left( {{c_{rk}}} \right)}} \right)}^2}} ,k \in \{ 1, \cdots ,{n_r}\} ,
\end{equation}
where $\left| {{I_r}} \right|$ represents the number of raw point reflectance labels with values greater than zero, and $r_{preds}^{\left( {{c_{rk}}} \right)}$ denotes the predicted reflectance for raw points.

\begin{figure}[!htb]
\begin{center}
    \begin{tabular}{c@{\hspace{-5mm}}}
    \hspace{-15pt}\includegraphics[width=3.5in]{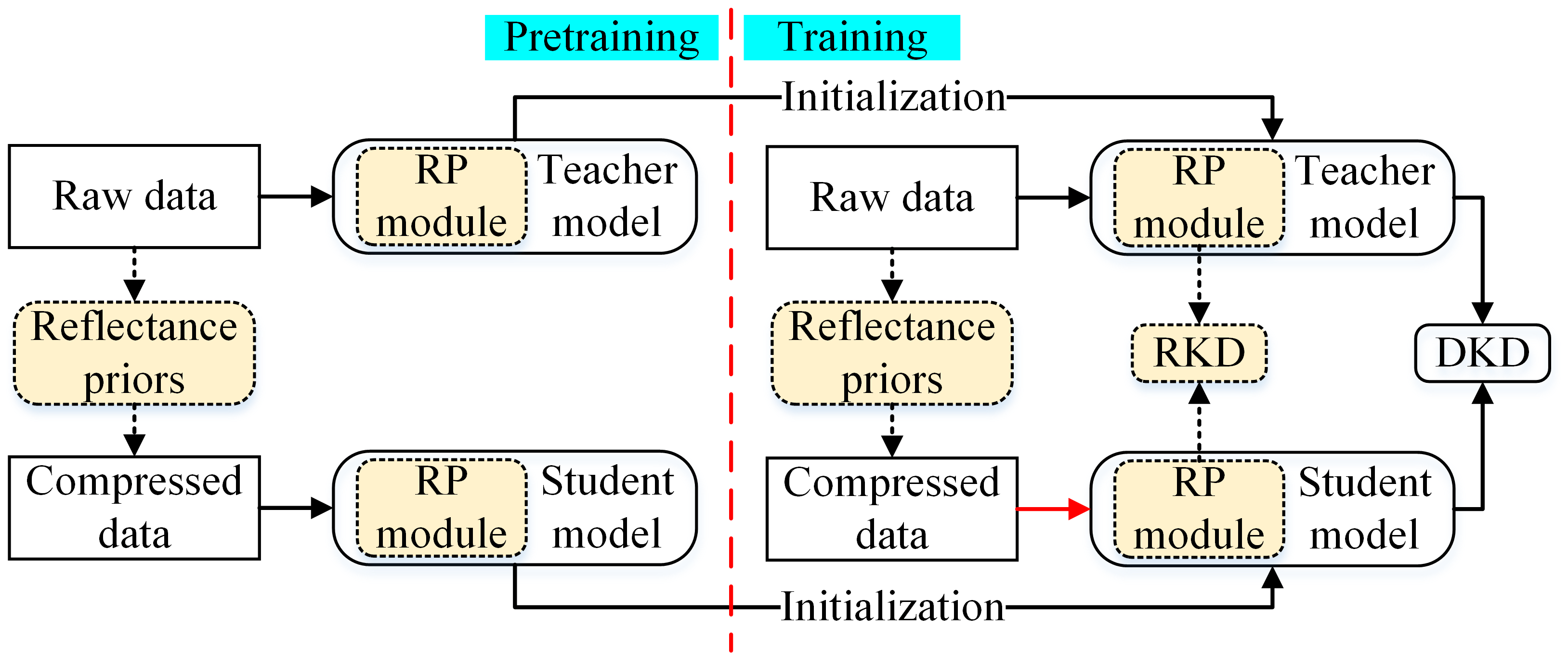}   \\
\end{tabular}
  \caption{\revise{Principle of CDTS. Dotted lines indicate specialized designs for compressed point-cloud reflectance prediction in RPKD (CDTS), while the red line represents the pipeline for evaluating final performance.}}
  \label{fig:figure5}
\vskip -10pt
\end{center}
\end{figure}

\subsection{\revise{Reflectance Prediction-based CDTS}}

\revise{Inspired by weak-data robustness research \cite{wei2022lidar,shan2023focal,huang2024sunshine}, the proposed CDTS employs raw and compressed data models in a teacher–student framework for knowledge distillation (as shown in Fig. \ref{fig:figure5}), thereby improving detection performance on low-quality compressed data. To tackle the critical challenge of missing LiDAR reflectance in lossy geometric compression, our RPKD (CDTS) strengthens reflectance reconstruction through the integration of reflectance priors, the RP module, and the RKD constraint.} In knowledge distillation, we introduce two components: RKD, which transfers reflectance prediction knowledge from the RP module, and DKD, which refines first-stage proposals in the two-stage 3D object detector. Both methods belong to response-based knowledge distillation, and their principles are outlined below.

\textbf{Reflectance Knowledge Distillation.} Building upon the RP module’s outputs, we obtain the predicted raw and compressed point reflectance.
Since they do not have a one-to-one correspondence, we map the recovered teacher reflectance from raw point-cloud voxels 
to corresponding compressed points. The mapped teacher reflectance $r_{\scriptscriptstyle RKD}^{\left( {{c_{ci}}} \right)}$ of each compressed point serves as a soft label of RKD. We also adopt MSELoss to constrain the reflectance distillation process, and the RKD loss function ${{\cal L}_{\scriptscriptstyle RKD}}$ is formulated as 
\begin{equation}
\label{9}
\begin{array}{*{20}{l}}
{{{\cal L}_{\scriptscriptstyle RKD}} = \frac{\textstyle 1}{{n_{fg}^{\left( {{{\cal P}_c}} \right)}}}\sum\limits_{i = 1}^{{n_{\scriptscriptstyle RKD}}} {{\cal F}{\cal G}\left( {{{\left( {r_{preds}^{\left( {{c_{ci}}} \right)} - r_{\scriptscriptstyle RKD}^{\left( {{c_{ci}}} \right)}} \right)}^2}} \right)} ,}\\
{i \in \{ 1, \cdots ,{n_{\scriptscriptstyle RKD}}\} ,}
\end{array}
\end{equation}
where ${\cal F}{\cal G}\left(  \cdot  \right)$ denotes screening foreground compression points during loss computation, ${n_{\scriptscriptstyle RKD}}$ is the number of compressed points that can be successfully matched with raw point-cloud voxels, and $n_{fg}^{\left( {{{\cal P}_c}} \right)}$ is the number of foreground compression points.

\textbf{Detection Knowledge Distillation.} In Fig. \ref{fig:figure3}, the two-stage 3D detector transfers key features from the first-stage network to the second-stage detector for detection refinement. Additionally, the first-stage dense proposals passed to the second stage inherently carry valuable detection knowledge for distillation. DKD combines the raw and compressed point-cloud proposals to form a teacher-student pair for response-based knowledge distillation.
For distillation constraints, we apply Logit KD proposed in SparseKD \cite{yang2022towards} to design the distillation loss. Therefore, the DKD loss function ${{\cal L}_{\scriptscriptstyle DKD}}$ can be expressed as:
\begin{equation}
\label{10}
\begin{array}{*{20}{l}}
{{{\cal L}_{\scriptscriptstyle DKD}} = {\alpha _1}{\cal L}_{cls}^{\scriptscriptstyle KD} + {\alpha _2}{\cal L}_{box}^{\scriptscriptstyle KD},}\\
{{\cal L}_{cls}^{\scriptscriptstyle KD} = \frac{\textstyle 1}{{{n_{fg}}}}\sum\limits_{i = 1}^{{n_{fg}}} {{\cal L}_{\scriptscriptstyle MSE}^{\left( i \right)}\left( {p_{cls}^{\left( {{{\cal P}_c}} \right)},p_{cls}^{\left( {{{\cal P}_r}} \right)}} \right)} ,}\\
{{\cal L}_{box}^{\scriptscriptstyle KD} = \frac{\textstyle 1}{{{n_{fg}}}}\sum\limits_{i = 1}^{{n_{fg}}} {{\cal L}_{\scriptscriptstyle SL}^{\left( i \right)}\left( {b_{box}^{\left( {{{\cal P}_c}} \right)},b_{box}^{\left( {{{\cal P}_r}} \right)}} \right)} ,}
\end{array}
\end{equation} 
where ${\cal L}_{\scriptscriptstyle MSE}^{\left( i \right)}$ refers to the MSELoss of classification scores ${p_{cls}}$, ${\cal L}_{\scriptscriptstyle SL}^{\left( i \right)}$ represents the Smooth-L1 loss of bounding boxes ${b_{box}}$, and ${n_{fg}}$ is the number of foreground objects in the first-stage proposals.

\subsection{Loss Function}
In this study, we pretrain separate detectors on compressed and raw point clouds and initialize the RPKD framework parameters with these pretrained models. Their loss functions are defined as follows:

\begin{equation}
\label{11}
\begin{array}{l}
{\cal L} = {\lambda _1}{\cal L}_{rp} + {\cal L}_{rpn} + {\cal L}_{rcnn},\\
{\cal L}_{rpn} = {\beta _1}{\cal L}_{cls} + {\beta _2}{\cal L}_{box} + {\beta _3}{\cal L}_{dir},
\end{array}
\end{equation}
where ${{\cal L}_{rp}}$, ${{\cal L}_{cls}}$, ${{\cal L}_{box}}$ and ${{\cal L}_{dir}}$ represent the reflectance prediction, classification, bounding box and orientation losses of the first-stage detector, respectively. The corresponding weight coefficients are ${\lambda _1}$, ${\beta _1}$, ${\beta _2}$ and ${\beta _3}$. ${{\cal L}_{rcnn}}$ denotes the second-stage detector loss with a weight coefficient of 1. On this basis, the overall loss function of our RPKD framework ${{\cal L}_{\scriptscriptstyle RPKD}}$ is formulated as follows:
\begin{equation}
\label{13}
\begin{array}{*{20}{l}}
{{{\cal L}_{\scriptscriptstyle RPKD}} = {\lambda _1}{\cal L}_{rp}^{\left( {{{\cal P}_c}} \right)} + \widehat {\cal L}_{rpn}^{\left( {{{\cal P}_c}} \right)} + {\cal L}_{rcnn} + {\lambda _2}{{\cal L}_{\scriptscriptstyle RKD}} + {\lambda _3}{{\cal L}_{\scriptscriptstyle DKD}},}\\
{{{\widehat {\cal L}}_{rpn}^{\left( {{{\cal P}_c}} \right)}} = {\beta _1}\widehat {\cal L}_{cls}^{\left( {{{\cal P}_c}} \right)} + {\beta _2}\widehat {\cal L}_{box}^{\left( {{{\cal P}_c}} \right)} + {\beta _3}\widehat {\cal L}_{dir}^{\left( {{{\cal P}_c}} \right)},}
\end{array}
\end{equation}
where $\widehat {\cal L}_{cls}^{\left( {{{\cal P}_c}} \right)}$, $\widehat {\cal L}_{box}^{\left( {{{\cal P}_c}} \right)}$ and $\widehat {\cal L}_{dir}^{\left( {{{\cal P}_c}} \right)}$ are the classification, bounding box and orientation losses of the first-stage detector based on Label KD in SparseKD \cite{yang2022towards}. The weight coefficients for the distillation losses are ${\lambda _2}$ and ${\lambda _3}$.

\section{Experiments}
\label{sec:experiments}
For 3D object detection in compressed point clouds, we employed LiDAR data from \revise{the KITTI and DAIR-V2X-V datasets}
as raw point-cloud inputs for the compression network. The geometric coordinates of these point clouds were compressed and reconstructed at multiple code rates using the voxel-based lossy geometric compression method PCC-S \cite{chen2022point}. \revise{Building on the proposed CDTS, extensive experiments demonstrate that our RPKD (CDTS) significantly enhances detection robustness for non-reflectance compressed point clouds.} The following sections detail the implementation, comparative results, and ablation analysis.

\begin{table*}[!t]
\centering
\caption{\revise{3D detection performance on the KITTI validation split, including AP and mAP with 40 recall positions. Bolded values indicate the highest performance in each comparison pair. Building on raw and non-reflectance compressed data, STS-C refers to the compressed-data single-source training strategy, while CDTS represents our cross-source distillation training strategy. *: Reimplementation results using public code on raw data.}}
\label{table:table1}
\begin{tabular}{c|c|c|ccc|ccc|ccc|c}
\hline
\multirow{2}{*}{Data Type} & \multirow{2}{*}{Bpp} & \multirow{2}{*}{Method} & \multicolumn{3}{c|}{Car 3D} & \multicolumn{3}{c|}{Ped. 3D} & \multicolumn{3}{c|}{Cyc. 3D} & \multirow{2}{*}{mAP} \\
& & & {Easy} & {Mod.} & {Hard} & {Easy} & {Mod.} & {Hard} & {Easy} & {Mod.} & {Hard} & \\
\hline
\hline
Raw Data & - & PV-RCNN \cite{shi2020pv}*   & 91.97    & 84.72   & 82.32   & 68.60    & 60.61    & 55.55   & 92.06    & 74.48    & 69.63   & 75.55 
\\
\hline
\multirow{2}{*}{PCC-S-C12} & \multirow{2}{*}{3.86} & PV-RCNN (STS-C)   & 91.48    & 82.51   & \textbf{80.72}   & 56.17    & 49.66    & 45.37   & 87.85    & 68.51    & \textbf{64.21}   & 69.61  \\
& & RPKD-PV (CDTS)  & \textbf{92.02}    & \textbf{83.13}   & 80.21   & \textbf{65.72}    & \textbf{58.68}    & \textbf{53.72}   & \textbf{90.39}    & \textbf{68.63}    & 63.92   & \textbf{72.94}   \\
\hline
\multirow{2}{*}{PCC-S-C11} & \multirow{2}{*}{2.15} & PV-RCNN (STS-C)   & 90.85    & 82.25   & \textbf{80.33}   & 57.79    & 50.83    & 46.20   & 88.92    & 65.79    & 61.40   & 69.37   \\
& & RPKD-PV (CDTS)   & \textbf{91.38}    & \textbf{82.95}   & 79.96   & \textbf{70.93}    & \textbf{62.81}    & \textbf{56.39}   & \textbf{89.21}    & \textbf{66.68}    & \textbf{62.09}   & \textbf{73.60}    \\
\hline
\multirow{2}{*}{PCC-S-C10} & \multirow{2}{*}{1.05} & PV-RCNN (STS-C) & 90.74    & 81.40   & 79.24   & 54.71    & 47.58    & 44.12   & 81.08    & 61.64    & 57.45   & 66.44  \\
& & RPKD-PV (CDTS)   & \textbf{91.53}    & \textbf{82.52}   & \textbf{79.45}   & \textbf{63.24}    & \textbf{57.24}    & \textbf{51.90}   & \textbf{85.51}    & \textbf{61.70}    & \textbf{57.68}   & \textbf{70.09}    \\
\hline
\hline
Raw Data & - & Voxel-RCNN \cite{deng2021voxel}* & 92.55    & 84.65   & 82.41   & 66.07    & 59.30    & 53.73   & 92.81    & 73.55    & 68.97   & 74.89   \\
\hline
\multirow{2}{*}{PCC-S-C12} & \multirow{2}{*}{3.86} & Voxel-RCNN (STS-C)  & \textbf{92.27}    & 82.79   & \textbf{80.17}   & 63.64    & 56.21    & 51.21   & 90.69    & 71.35    & \textbf{66.92}   & 72.81  \\
& & RPKD-Voxel (CDTS)   & 91.89    & \textbf{82.94}   & 80.03   & \textbf{71.49}    & \textbf{64.33}    & \textbf{58.50}   & \textbf{92.28}    & \textbf{71.47}    & 66.85   & \textbf{75.53}  \\
\hline
\multirow{2}{*}{PCC-S-C11} & \multirow{2}{*}{2.15} & Voxel-RCNN (STS-C)   & \textbf{92.03}    & 82.72   & 80.16   & 61.49    & 54.77    & 49.30   & 88.27    & 67.99    & \textbf{63.97}   & 71.19   \\
& & RPKD-Voxel (CDTS)   & 91.74    & \textbf{82.90}   & \textbf{80.22}   & \textbf{69.65}    & \textbf{62.83}    & \textbf{56.10}   & \textbf{91.80}    & \textbf{68.36}    & 63.94   & \textbf{74.17}  \\
\hline
\multirow{2}{*}{PCC-S-C10} & \multirow{2}{*}{1.05} & Voxel-RCNN (STS-C)   & 90.96    & 81.37   & 78.73   & 60.10    & 51.66    & 45.88   & 83.05    & 62.86    & 58.62   & 68.14   \\
& & RPKD-Voxel (CDTS)  & \textbf{91.37}    & \textbf{82.33}   & \textbf{79.50}   & \textbf{66.72}    & \textbf{59.53}    & \textbf{53.88}   & \textbf{84.68}    & \textbf{63.03}    & \textbf{58.93}   & \textbf{71.11}   \\
\hline
\hline
Raw Data & - & SECOND \cite{yan2018second}* & 90.43 & 81.50 & 78.50 & 55.89 & 50.97 & 46.38 & 81.91 & 64.40 & 60.53 & 67.83  \\
\hline
\multirow{2}{*}{PCC-S-C12} & \multirow{2}{*}{3.86} & SECOND (STS-C) & 90.34 & 81.21 & 78.32 & 55.59 & 49.67 & 45.32 & 80.19 & 65.19 & 61.23 & 67.45 \\
& & RPKD-SECOND (CDTS) & \textbf{91.08} & \textbf{81.96} & \textbf{78.87} & \textbf{58.42} & \textbf{53.06} & \textbf{48.55} & \textbf{86.00} & \textbf{68.38} & \textbf{64.35} & \textbf{70.07} \\
\hline
\multirow{2}{*}{PCC-S-C11} & \multirow{2}{*}{2.15} & SECOND (STS-C) & 90.17 & 80.90 & 78.09 & 52.58 & 47.66 & 43.16 & 81.92 & 61.21 & 57.31 & 65.89 \\
& & RPKD-SECOND (CDTS) & \textbf{90.87} & \textbf{81.69} & \textbf{78.62} & \textbf{57.35} & \textbf{52.88} & \textbf{48.44} & \textbf{84.84} & \textbf{66.07} & \textbf{61.79} & \textbf{69.17} \\
\hline
\multirow{2}{*}{PCC-S-C10} & \multirow{2}{*}{1.05} & SECOND (STS-C) & 88.25 & 78.80 & 75.72 & 50.48 & 45.29 & 41.37 & 75.59 & 56.28 & 52.66 & 62.72 \\
& & RPKD-SECOND (CDTS) & \textbf{90.08} & \textbf{80.74} & \textbf{77.52} & \textbf{56.76} & \textbf{52.34} & \textbf{47.71} & \textbf{81.81} & \textbf{62.39} & \textbf{58.32} & \textbf{67.52} \\
\hline
\end{tabular}
\end{table*}

\subsection{Implementation Details}
\label{section:4A}
\subsubsection{Datasets and Hardware}
In KITTI, we selected 7,481 labeled training samples as raw point clouds, dividing them into a training split of 3,712 samples and a validation split of 3,769 samples. \revise{Similarly, in DAIR-V2X-V, we selected 6,509 training frames, which were split into 4,335 frames for training and 2,174 frames for validation.} We trained the PCC-S network on the raw point-cloud training split for compression learning and applied to reconstruct point clouds at octree levels 12, 11, and 10, denoted as PCC-S-C12, PCC-S-C11, and PCC-S-C10, respectively. The compression network was optimized for 20 epochs on a single RTX 8000 GPU with a batch size of 2 and a maximum octree level of 12. Independent detection models for each code rate were trained for 80 epochs on 4 RTX 4090 GPUs with a batch size of 16.


\subsubsection{Training Parameters}
In the RCM module, we matched compressed points with raw point-cloud voxels within a 1-voxel matching range, corresponding to $3\rm{\times}3\rm{\times}3$ spatial voxels. During pretraining, the raw and compressed point-cloud detectors shared the same loss weights, and these loss weights $\left( {{\lambda _1},{\beta _1},{\beta _2},{\beta _3}} \right)$ were set to $\left( {3,1,2,0.2} \right)$. For CDTS, we introduced two distillation methods, namely RKD and DKD, whose loss weights $\left( {{\lambda _2},{\lambda _3}} \right)$ were configured as $\left( {10,1} \right)$. Specifically, the distillation constraints for DKD included classification distillation loss ${\cal L}_{cls}^{\scriptscriptstyle KD}$ and bounding box distillation loss ${\cal L}_{box}^{\scriptscriptstyle KD}$, and their loss weights $\left( {{\alpha _1},{\alpha _2}} \right)$ were $\left( {6,0.5} \right)$. 

\subsubsection{Evaluation Metrics}
\revise{We evaluated the robustness of our RPKD framework for 3D object detection on compressed point clouds using the KITTI and DAIR-V2X-V validation splits. For KITTI and DAIR-V2X-V, }the primary metric is average precision (AP) at 40 recall positions with 3D IoU thresholds of 0.7, 0.5, and 0.5 for car, pedestrian, and cyclist categories, respectively. 
The Bit-per-point (Bpp) metric represents the average number of bits required to encode each point, indicating the code rate of compressed point clouds.

\begin{figure}[!b]
\centering
\setlength{\tabcolsep}{0pt}
\renewcommand{\arraystretch}{0}
\begin{tabular}{cc}
\subfloat[]{\includegraphics[width=1.75in]{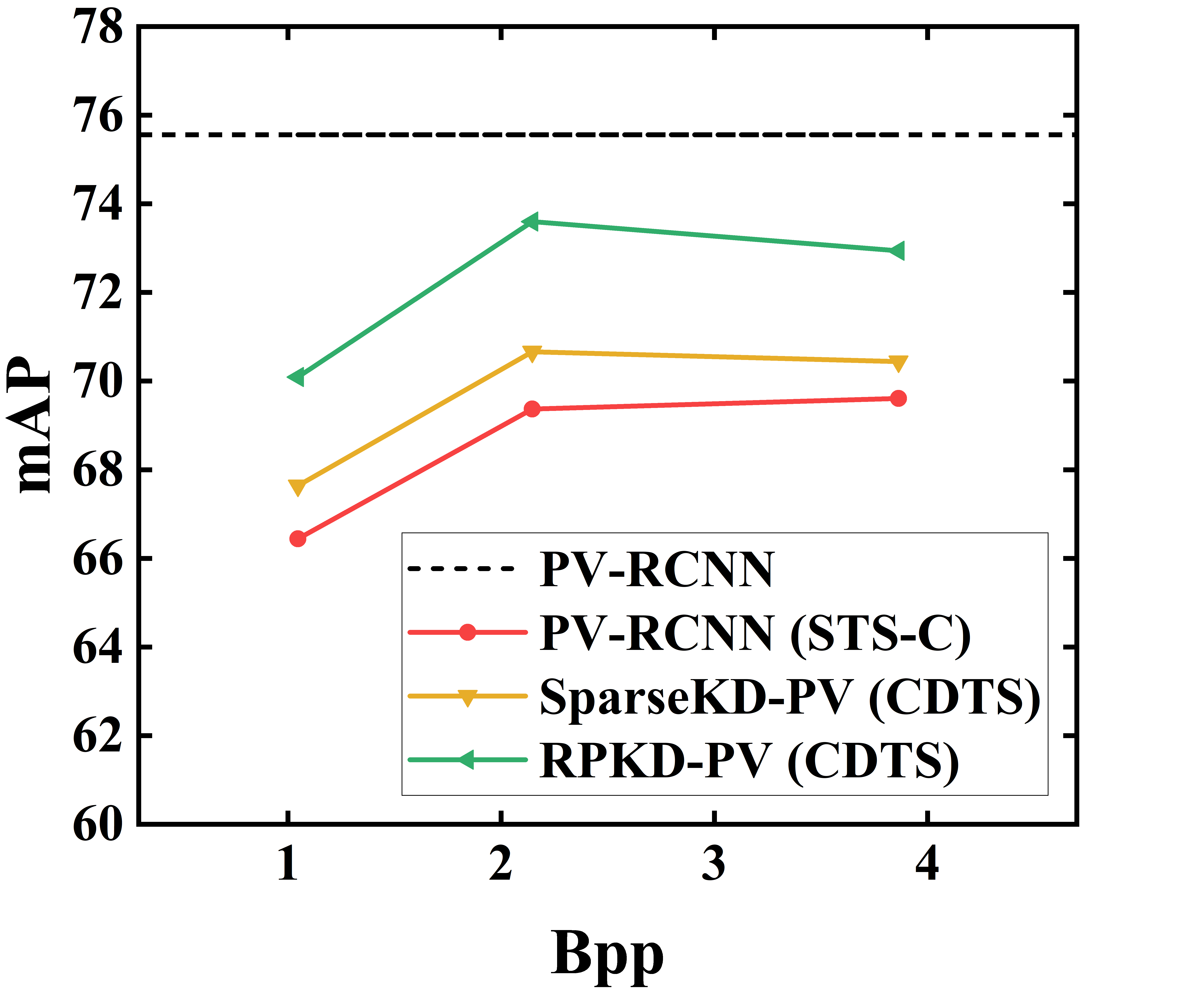}\label{fig:6a}} &
\subfloat[]{\includegraphics[width=1.75in]{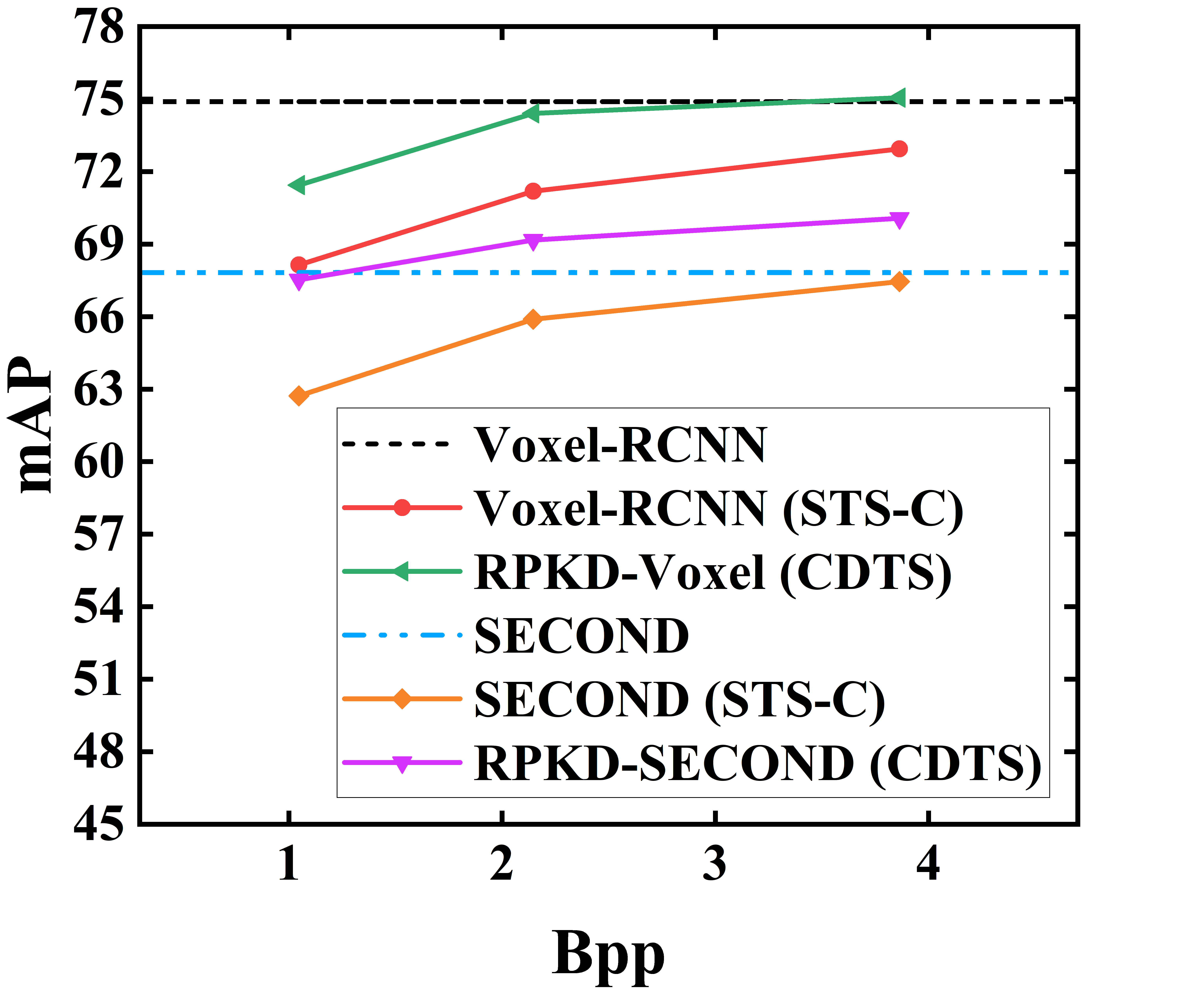}\label{fig:6b}}
\end{tabular}
\caption{\revise{Quantitative results of 3D object detection on the KITTI validation split. (a) Comparisons of different training strategies and distillation methods under the PV-RCNN baseline. (b) Generalization verification of our RPKD framework on the Voxel-RCNN and SECOND backbones. 
}}
\label{fig:figure6}
\end{figure}

\begin{table*}[!t]
\centering
\caption{\revise{3D detection performance on the DAIR-V2X-V validation split, including AP and mAP with 40 recall positions. Bolded values indicate the highest performance in each comparison pair. *: Reimplementation results using public code on raw data.}
}
\label{table:tabledair}
\begin{tabular}{c|c|c|ccc|ccc|ccc|c}
\hline
\multirow{2}{*}{Data Type} & \multirow{2}{*}{Bpp} & \multirow{2}{*}{Method} & \multicolumn{3}{c|}{Car 3D} & \multicolumn{3}{c|}{Ped. 3D} & \multicolumn{3}{c|}{Cyc. 3D} & \multirow{2}{*}{mAP} \\
& & & {Easy} & {Mod.} & {Hard} & {Easy} & {Mod.} & {Hard} & {Easy} & {Mod.} & {Hard} & \\
\hline
\hline
Raw Data & - & PV-RCNN \cite{shi2020pv}*  & 72.06 & 60.50 & 57.58 & 42.67 & 41.94 & 41.93 & 41.12 & 38.64 & 39.22 & 48.41 \\
\hline
\multirow{2}{*}{PCC-S-C12} & \multirow{2}{*}{4.01} & PV-RCNN (STS-C) & 71.76 & 60.15 & 57.22 & 37.65 & 36.14 & 36.09 & 34.57 & 34.24 & 34.65 & 44.72 \\
& & RPKD-PV (CDTS) & \textbf{72.79} & \textbf{61.08} & \textbf{58.16} & \textbf{41.97} & \textbf{41.51} & \textbf{40.77} & \textbf{45.53} & \textbf{41.53} & \textbf{42.10} & \textbf{49.49} \\
\hline
\multirow{2}{*}{PCC-S-C11} & \multirow{2}{*}{2.13} & PV-RCNN (STS-C) & 70.49 & 58.79 & 55.48 & 30.79 & 30.98 & 30.85 & 34.00 & 32.09 & 32.42 & 41.77 \\
& & RPKD-PV (CDTS) & \textbf{70.73} & \textbf{58.96} & \textbf{56.03} & \textbf{36.42} & \textbf{35.58} & \textbf{34.59} & \textbf{40.06} & \textbf{36.60} & \textbf{36.99} & \textbf{45.11} \\
\hline
\multirow{2}{*}{PCC-S-C10} & \multirow{2}{*}{0.96} & PV-RCNN (STS-C) & \textbf{68.13} & 56.49 & 52.06 & 8.52 & 8.77 & 8.96 & 12.86 & 13.18 & 12.43 & 26.82 \\
& & RPKD-PV (CDTS) & 68.03 & \textbf{57.11} & \textbf{52.69} & \textbf{10.01} & \textbf{10.35} & \textbf{10.29} & \textbf{14.98} & \textbf{14.88} & \textbf{14.54} & \textbf{28.10} \\
\hline
\end{tabular}
\end{table*}

\subsection{Experimental Results}
\label{section:4B}
Table \ref{table:table1} presents the performance improvement of our RPKD framework on compressed point clouds, \revise{using the PV-RCNN, Voxel-RCNN, and SECOND backbones on the KITTI validation split. For PCC-S-C12, PCC-S-C11, and PCC-S-C10 compressed point clouds, RPKD-PV increases mAP by 3.33, 4.23, and 3.65, respectively; RPKD-Voxel improves mAP by 2.72, 2.98, and 2.97; and RPKD-SECOND achieves mAP gains of 2.62, 3.28, and 4.8.} As a result, it reveals that our RPKD framework's detectors exhibit consistent effectiveness and robustness across various code rates and backbones. Additionally, for compressed point clouds with the same code-rate order, RPKD-PV increases the AP values for moderate-level pedestrians by 9.02, 11.98, and 9.66, respectively; \revise{RPKD-Voxel boosts the AP values by 8.12, 8.06, and 7.87; and RPKD-SECOND improves the AP values by 3.39, 5.22, and 7.05.} These findings highlight the remarkable enhancement in 3D detector robustness for challenging small-sized pedestrians in compressed point clouds.

\begin{table}[t]
\centering
\caption{3D detection performance of different RP module settings on the KITTI PCC-S-C11 validation split. The results are obtained by training on the 15\% KITTI PCC-S-C11 training split. The bolded value indicates the highest performance.}
\label{table:table3}
\setlength\tabcolsep{4pt}
\begin{tabular}{cc|cc|cc|c}
\hline
\multicolumn{2}{c|}{${\lambda _1}$} & \multicolumn{2}{c|}{$\left[ {{\rm{Ks,Pd,}}\left( {{\rm{Rd1,Rd2}}} \right)} \right]$}  & \multicolumn{2}{c|}{Layer Quantity} & \multirow{2}{*}{mAP} \\
1 & 3 & $\left[ {3,1,\left( {0.4,0.8} \right)} \right]$ & $\left[ {5,2,\left( {0.8,1.2} \right)} \right]$ & ${1 \times}$ & $\left( {1 \times ,2 \times } \right)$ & \\
\hline
\hline
${\rm{\times}}$ & $\surd$  & ${\rm{\times}}$  & $\surd$  & $\surd$  & ${\rm{\times}}$  & 61.31   \\
${\rm{\times}}$  & $\surd$  & $\surd$  & ${\rm{\times}}$  & ${\rm{\times}}$  & $\surd$  & 61.81   \\
$\surd$  & ${\rm{\times}}$  & $\surd$  & ${\rm{\times}}$  & $\surd$  & ${\rm{\times}}$  & 62.59   \\
${\rm{\times}}$ & $\surd$  & $\surd$  & ${\rm{\times}}$  & $\surd$  & ${\rm{\times}}$  & \textbf{63.70}   \\
\hline
\end{tabular}
\end{table}

As depicted in Fig. \hyperref[fig:figure6]{\ref{fig:figure6}(a)}, we compared the detection performance of PV-RCNN (STS-C), SparseKD-PV (CDTS), and RPKD-PV (CDTS) on the KITTI compressed data across multiple code rates, using the PV-RCNN backbone. RPKD-PV outperforms SparseKD-PV at all code rates, achieving the highest mAP values.
\revise{Fig. \hyperref[fig:figure6]{\ref{fig:figure6}(b)} shows the generalization performance of our RPKD framework on the Voxel-RCNN and SECOND backbones. Both RPKD-Voxel (CDTS) and RPKD-SECOND (CDTS) surpass their respective baselines across all code rates, with the two-stage RPKD-Voxel achieving higher mAP than the single-stage RPKD-SECOND.} These results vividly illustrate the exceptional robustness and significant performance improvements of our method for 3D object detection on compressed point clouds.

\revise{To assess the generalization ability of our RPKD framework, we evaluated its compressed data detection performance with the PV-RCNN backbone on DAIR-V2X-V. Table \ref{table:table3} further shows that RPKD-PV increases mAP by 4.77, 3.34, and 1.28 on the DAIR-V2X-V validation split.} These experimental results demonstrate that our RPKD framework consistently delivers strong performance across varying code rates, backbone networks, and detection datasets.

\begin{table}[!t]
\centering
\caption{3D detection performance of critical elements in our RPKD framework on the KITTI PCC-S-C11 validation split, where ‘w/’ and ‘w/o’ denote ‘with’ and ‘without’, respectively. The bolded value indicates the highest performance.}
\label{table:table4}
\setlength\tabcolsep{4pt}
\begin{tabular}{c|ccc|cc|c}
\hline
\multirow{2}{*}{${{\cal M}_c}\left(  \cdot  \right)$} & \multicolumn{3}{c|}{Options of RKD} & \multicolumn{2}{c|}{Detectors of DKD} & \multirow{2}{*}{mAP} \\
& ${\cal F}{\cal G}\left(  \cdot  \right)$ & Voxel-wise & Point-wise & w/o RP & w/ RP & \\
\hline
\hline
${\rm{\times}}$ & ${\rm{\times}}$  & ${\rm{\times}}$  & ${\rm{\times}}$  & ${\rm{\times}}$  & ${\rm{\times}}$  & 71.39  \\
$\surd$ & ${\rm{\times}}$  & ${\rm{\times}}$  & ${\rm{\times}}$  & ${\rm{\times}}$  & ${\rm{\times}}$  & 72.00  \\
${\rm{\times}}$ & $\surd$  & ${\rm{\times}}$  & $\surd$  & ${\rm{\times}}$  & ${\rm{\times}}$  & 69.12  \\
$\surd$ & ${\rm{\times}}$  & $\surd$ & ${\rm{\times}}$  & ${\rm{\times}}$  & ${\rm{\times}}$  & 70.16  \\
$\surd$ & $\surd$  & $\surd$  & ${\rm{\times}}$  & ${\rm{\times}}$  & ${\rm{\times}}$ & 71.68  \\
$\surd$ & $\surd$  & $\surd$  & ${\rm{\times}}$  & $\surd$  & ${\rm{\times}}$  & 71.57  \\
$\surd$ & $\surd$  & $\surd$  & ${\rm{\times}}$  & ${\rm{\times}}$  & $\surd$  & \textbf{73.60}  \\
\hline
\end{tabular}
\end{table}

\begin{table}[!t]
\centering
\caption{3D detection performance of the designed modules in our RPKD framework on the KITTI PCC-S-C11 validation split. The bolded value indicates the highest performance.}
\label{table:table5}
\begin{tabular}{ccc|c}
\hline
RP   Module & RKD & DKD & mAP   \\
\hline
\hline
${\rm{\times}}$  & ${\rm{\times}}$   & ${\rm{\times}}$   & 69.37 \\
$\surd$  & ${\rm{\times}}$   & ${\rm{\times}}$   & 72.00 \\
$\surd$  & $\surd$   & ${\rm{\times}}$   & 71.68 \\
$\surd$  & $\surd$   & $\surd$   & \textbf{73.60} \\
\hline
\end{tabular}
\end{table}

\subsection {Ablation Study}
\label{section:4C}
We conducted comprehensive ablation experiments with the PV-RCNN backbone to verify the effectiveness of our RPKD framework. In this section, we used raw point clouds from the KITTI dataset and their corresponding compressed point clouds of PCC-S-C11 as experimental data. The evaluation metric is the mAP of compressed point-cloud detectors on the KITTI validation split.

\subsubsection{Different Configurations of RP Module}
For non-reflectance compressed point-cloud voxels, the RP module employed 3D sparse convolution and voxel set extraction to generate point-wise geometric features for compressed point reflectance prediction. In terms of expanding receptive field, we modified the kernel size $\left( {{\rm{Ks}}} \right)$ and padding $\left( {{\rm{Pd}}} \right)$ of ${1 \times}$ 3D sparse convolution, as well as the pooling radius $\left( {{\rm{Rd1, Rd2}}} \right)$ of ${1 \times}$ voxel set extraction. To assess the impact of network layer quantity on detection performance, we designed two network schemes: a single-layer network ${1 \times}$ and a double-layer network $\left( {1 \times ,2 \times } \right)$. Additionally, we also adjusted the RP module’s loss weight for performance comparison. As shown in Table \ref{table:table3}, when using the single-layer network ${1 \times}$ with $\left[ {{\rm{Ks,Pd,}}\left( {{\rm{Rd1,Rd2}}} \right)} \right] = \left[ {3,1,\left( {0.4,0.8} \right)} \right]$ and ${\lambda _1} = 3$, the compressed point-cloud detector achieves optimal detection performance.

\subsubsection{Critical Elements of Our RPKD}
As shown in Table \ref{table:table4}, we evaluated the impact of critical elements in our RPKD framework on the compressed data detection performance. In the first two rows, we compared the pretraining performance of compressed point-cloud detectors, where ${{\cal M}_c}\left(  \cdot  \right)$ refers to the voxel-based averaging operation in the RCM module. ${\cal F}{\cal G}\left(  \cdot  \right)$ denotes the process of screening foreground compressed points for RKD. In the third row, we used the matched raw point reflectance as the distillation label during point-wise RKD, without performing the voxel-based ${{\cal M}_c}\left(  \cdot  \right)$. In contrast, the fourth row employed the matched reflectance of raw point-cloud voxels as the distillation label during voxel-wise RKD, with the voxel-based ${{\cal M}_c}\left(  \cdot  \right)$. The sixth and final rows assessed the effect of incorporating the RP mechanism into the teacher model, which provides the first-stage proposals for DKD. As indicated in the last row, our RPKD framework achieves the best performance under these settings.

\subsubsection{Designed Modules of Our RPKD}
As shown in Table \ref{table:table5}, we evaluated the designed module's effectiveness for 3D object detection on compressed point clouds. 
The first row reports the baseline performance of PV-RCNN (STS-C) on PCC-S-C11. We then incrementally integrated the RP module, RKD and DKD into the detector and recorded the corresponding mAP values. Each module contributes to performance improvement, with the highest mAP achieved when all modules are combined.

\begin{table}[!t]
\centering
\caption{\revise{Performance comparison of advanced one-stage and two-stage methods on the KITTI PCC-S-C11 validation split, including mAP with 40 recall positions for each category. Bolded values indicate the best performance. \vstarup: Reimplementation results using public code on compressed data of PCC-S-C11.}}
\label{table:table6}
\setlength{\tabcolsep}{2pt}
\begin{tabular}{c|c|ccc|c}
\hline
Method & Stage & Car 3D & Ped. 3D & Cyc. 3D & Overall \\
\hline
\hline
PointPillars (STS-C) \cite{lang2019pointpillars}\smash{\vstarup} & \multirow{4}{*}{One} & 80.45 & 45.22 & 64.76 & 63.48 \\
CenterPoint (STS-C) \cite{yin2021center}\smash{\vstarup} & & 80.87 & 50.74 & 65.88 & 65.83 \\
SECOND (STS-C) \cite{yan2018second}\smash{\vstarup} & & 83.06 & 47.80 & 66.81 & 65.89 \\
RPKD-SECOND (CDTS) & & 83.72 & 52.89 & 70.90 & 69.17 \\
\hline
PV-RCNN (STS-C) \cite{shi2020pv}\smash{\vstarup} & \multirow{5}{*}{Two} & 84.48 & 51.61 & 72.03 & 69.37 \\
PV-RCNN++ (STS-C) \cite{shi2023pv}\smash{\vstarup} & & 84.56 & 58.88 & 71.88 & 71.77 \\
Voxel-RCNN (STS-C) \cite{deng2021voxel}\smash{\vstarup} & & \textbf{84.97} & 55.19 & 73.41 & 71.19 \\
RPKD-PV (CDTS) & & 84.76 & \textbf{63.38} & 72.66 & 73.60 \\
RPKD-Voxel (CDTS) & & 84.95 & 62.86 & \textbf{74.70} & \textbf{74.17} \\
\hline
\end{tabular}
\end{table}

\subsubsection{Comparison of Single-Stage and Two-Stage Methods}
\revise{In Table \ref{table:table6}, we compared our method against advanced single-stage detectors (PointPillars, CenterPoint, and SECOND) and two-stage detectors (PV-RCNN, PV-RCNN++, and Voxel-RCNN) on PCC-S-C11. Our RPKD (CDTS) improves overall mAP by 3.28, 4.23, and 2.98 on the SECOND, PV-RCNN, and Voxel-RCNN backbones, respectively. Notably, RPKD-Voxel (CDTS) achieves the highest overall mAP across all detection methods. These results demonstrate that our method enhances detection performance for both single-stage and two-stage detectors, with two-stage models exhibiting stronger detection robustness.}

\begin{figure}[!t]
\begin{center}
    \begin{tabular}{c@{\hspace{-4mm}}}
    \hspace{-15pt}\includegraphics[width=3.3in]{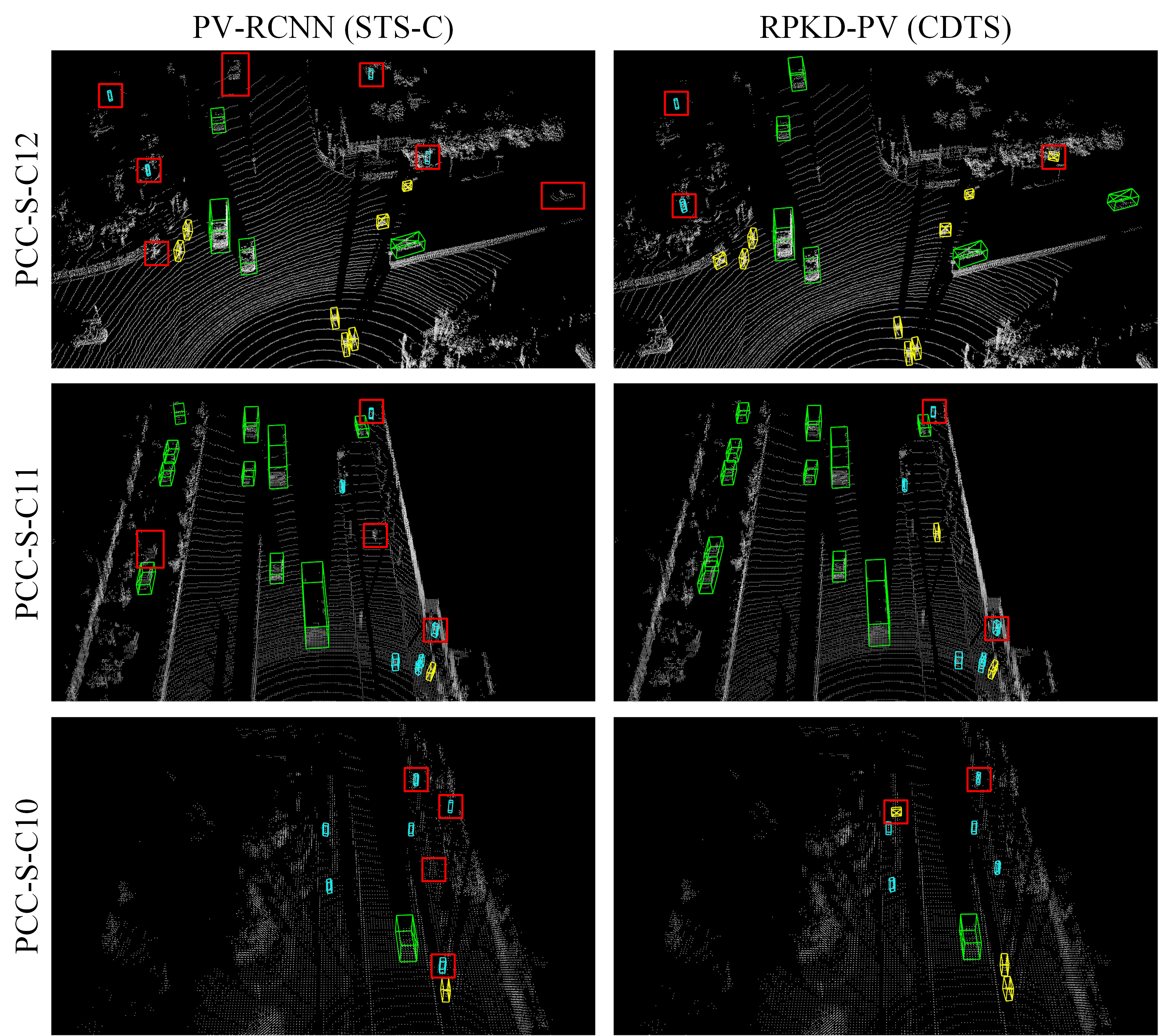}   \\
\end{tabular}
  \caption{\revise{Visual comparison of different detection methods on the DAIR-V2X-V validation split. The false positives and missed detections are highlighted with red rectangles.}}
  \label{fig:figure7}
    \vspace{-2em}
\end{center}
\end{figure}

\begin{figure*}[!htb]
\begin{center}
    \begin{tabular}{c@{\hspace{-4mm}}}
    \hspace{-15pt}\includegraphics[width=6in]{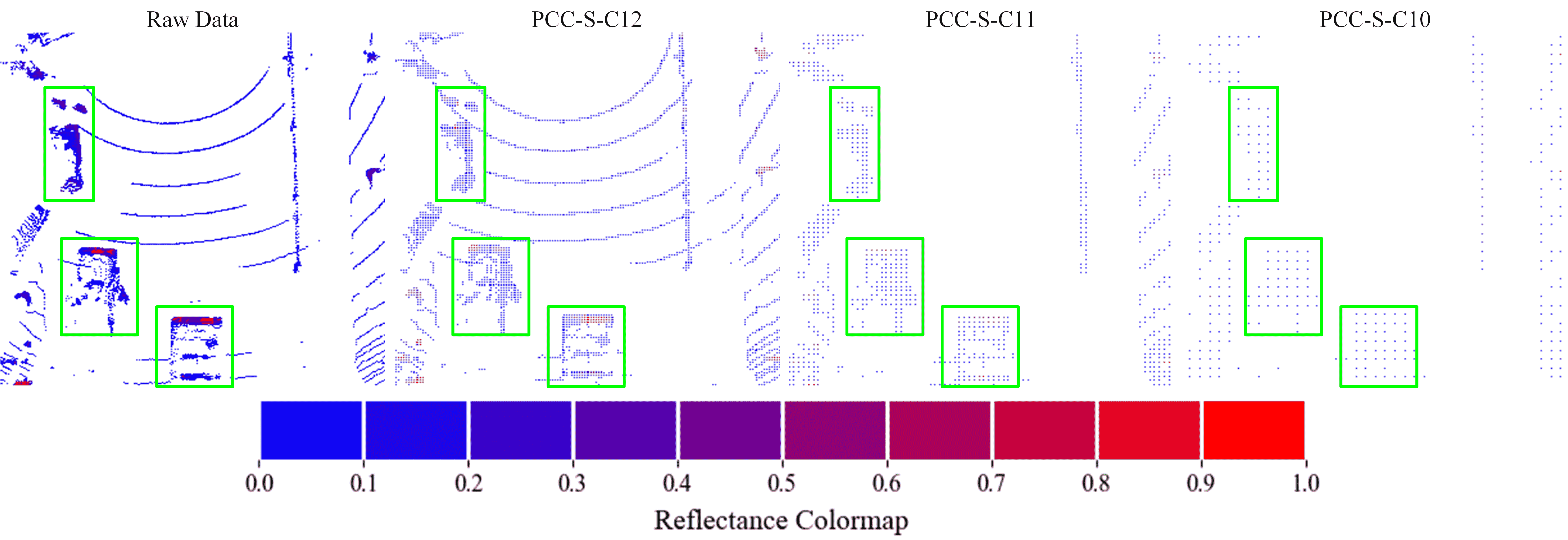}   \\
\end{tabular}
  \caption{\revise{Comparison of BEV reflectance visualization on the DAIR-V2X-V dataset: raw data colored by actual reflectance versus compressed point clouds colored by RP module predictions under different code rates. Green boxes highlight foreground objects for clarity.}}
  \label{fig:figure8}
\vskip -10pt
\end{center}
\end{figure*}

\subsubsection{\revise{Visual Results on DAIR-V2X-V}}
\revise{Fig. \ref{fig:figure7} presents a visual comparison on the DAIR-V2X-V validation split. RPKD-PV (CDTS) produces fewer false positives and missed detections across multiple code rates, confirming its robustness. As shown in Fig. \ref{fig:figure8}, we used the maximum reflectance value from either raw or compressed data to color the corresponding visual points within each BEV grid. For PCC-S-C12 and PCC-S-C11, the predicted reflectance from compressed data closely matches the actual reflectance from raw data in the primary reflective regions of foreground objects, highlighting the effectiveness of the RP module. However, for PCC-S-C10, the limited number and low density of compressed points lead to a reduction in accuracy, diminishing the reliability of the reflectance prediction.}



\subsection{\revise{Analysis of Compression Transmission}}
\revise{In large-scale V2X scenarios, early fusion is constrained by the high bandwidth required to transmit raw LiDAR data. To address this, we employed voxel-based lossy geometric compression and omitted reflectance encoding, thereby maximizing bandwidth savings. For compression transmission–based cooperative perception, our analysis and evaluation proceed as follows.}

\subsubsection{\revise{Memory and Bandwidth of Raw LiDAR Data}}
\revise{Onboard LiDAR sensors capture point cloud data consisting of single-precision floating-point coordinates (X, Y, Z) and 8-bit unsigned integer reflectance values. The memory usage of a coordinate and reflectance component is denoted as $\left( {{m_c},{m_r}} \right)$. Accordingly, the memory per raw point is ${m_0} = 3 \times {m_c} + {m_r}$. For LiDAR sensors in the KITTI and DAIR-V2X-V datasets, the number of points acquired per second is ${N_0}$. The corresponding transmission bandwidth is ${B_0} = {N_0} \times {m_0}$, with the detailed requirements summarized in Table \ref{table:table7}.}

\begin{table}[!t]
\centering
\caption{\revise{
Memory and bandwidth requirements for KITTI and DAIR-V2X-V raw data.}}
\label{table:table7}
\setlength{\tabcolsep}{2pt}
\begin{tabular}{c|c|c|c|c}
\hline
Dataset & Name  & Symbol & Value  & Unit \\
\hline
\hline
\multirow{3}{*}{\shortstack{Common \\Parameters}} & Single Coordinate Memory & $m_c$ & 32         & Bit  \\
& Reflectance Memory & $m_r$ & 8 & Bit  \\
& Single Point Memory & $m_0$ & 104 & Bit  \\
\hline
\multirow{2}{*}{KITTI} & Points Per Second & $N_0$ & $1.3 \times 10^6$ & - \\
& Transmission Bandwidth & $B_0$ & 135.2 & Mbps \\
\hline
\multirow{2}{*}{DAIR-V2X-V} & Points Per Second & $N_0$ & $7.2 \times 10^5$ & - \\
& Transmission Bandwidth & $B_0$ & 74.88 & Mbps \\
\hline
\end{tabular}
\end{table}

\revise{In vehicular networking, prior studies \cite{balti2021situational,govindarajulu2019range,ni2023cellfusion} report that the channel capacity for connected vehicles typically ranges from 30 Mbps to 1 Gbps under varying traffic conditions. Within these limits, transmitting raw point clouds across multiple vehicles overloads both uplink and downlink channels, making real-time data sharing impractical. This highlights the necessity of point cloud compression, where data are compressed and encoded before transmission to enable efficient and feasible cooperative perception.}

\subsubsection{\revise{Bandwidth Calculation for Compressed Point-Cloud Transmission}}
\revise{Existing LiDAR point cloud compression modes can be classified into three categories based on the degree of information loss: lossless compression, lossy compression, and lossy geometry compression. In G-PCC \cite{cao2021compression}, the number of points per second in losslessly compressed point clouds remains identical to that of the raw point clouds, denoted as $N_0$. The corresponding transmission bandwidth $B_{Lossless}$ is defined as:}
\begin{equation}
\label{1}
{B_{Lossless}} = {N_0} \cdot \left( {{t_c} + {t_r}} \right)
\end{equation}
\revise{where $t_c$ and $t_r$ denote the code rates for coordinates and reflectance in losslessly compressed point clouds, respectively. In G-PCC–based lossy compression, the number of compressed points decreases according to the octree depth. Let $N_1 < N_0$ denote the number of points per second in lossy compressed point clouds, thus the transmission bandwidth $B_{Lossy}$ is expressed as:}
\begin{equation}
\label{2}
{B_{Lossy}} = {N_1} \cdot \left( {{{\hat t}_c} + {{\hat t}_r}} \right)
\end{equation}
\revise{where ${\hat t_c}$ and ${\hat t_r}$ represent the code rates for coordinates and reflectance in lossy compressed point clouds, respectively. Moreover, lossy geometry compression further reduces transmission bandwidth by discarding reflectance encoding. Assuming the number of points per second remains $N_1$, the transmission bandwidth $B_{Lossy-G}$ for lossy geometry compressed point clouds is obtained by the following formula:}
\begin{equation}
\label{3}
{B_{Lossy - G}} = {N_1} \cdot {\hat t_c}
\end{equation}

\subsubsection{\revise{Evaluation for Multi-Vehicle Communication}}
\revise{Based on G-PCC, Table \ref{table:table8} reports the multi-vehicle communication performance of different compression modes on the KITTI dataset. The key metrics include coordinate Bpp, reflectance Bpp, transmission bandwidth, and channel margin. Since coordinate and reflectance code rates lack a fixed proportional relationship, we selected representative values from prior compression studies \cite{chen2022point, wang2024versatile2} under medium compression ratio settings, corresponding to a point retention rate of 0.85. The channel capacity $C$ for connected vehicles was set to 200 Mbps, a level achievable in most traffic scenarios. The channel margin $\hat C$ for a given vehicle is defined as:}
\begin{equation}
\label{4}
\hat C = C - \left( {{N_{car}} - 1} \right) \cdot \left( {{B_r} + {B_s}} \right)
\end{equation}
\revise{where ${B_r}$ and ${B_s}$ denote the bandwidth required for receiving and sending point clouds with a single neighboring vehicle, and ${N_{car}}$ is the number of connected vehicles, all adopting the same compression setting.}

\begin{table*}[!t]
\centering
\caption{\revise{
Multi-vehicle communication performance of different G-PCC compression modes on the KITTI dataset. The key metrics include coordinate Bpp, reflectance Bpp, transmission bandwidth, and channel margin.}}
\label{table:table8}
\setlength{\tabcolsep}{3pt}
\begin{tabular}{c|c|c|c|c|c|c|cc}
\hline
\multirow{2}{*}{\shortstack{Compression \\Mode}} & \multirow{2}{*}{\shortstack{Point Retention \\Rate}} & \multirow{2}{*}{\shortstack{Points Per \\Second}} & \multirow{2}{*}{\shortstack{Coordinate \\Bpp}} & \multirow{2}{*}{\shortstack{Reflectance \\Bpp}} & \multirow{2}{*}{\shortstack{Bandwidth$\downarrow$ \vspace{-0.1em} \\ (Mbps)} } & \multirow{2}{*}{\shortstack{Channel Capacity \vspace{-0.1em} \\ (Mbps)}} & \multicolumn{2}{c}{Channel Margin$\uparrow$ (Mbps)} \\
  &  &  &  &  &  &  & ${N_{car}} = 2$ & ${N_{car}} = 11$  \\
\hline
\hline
Uncompressed  & 1.0  & $1.3 \times 10^6$ & - & - & 135.2 & 200  & -70.4 & -2504.0 \\
Lossless & 1.0 & $1.3 \times 10^6$ & 20.22 & 4.76 & 32.47 & 200 & 135.06 & -449.4 \\
Lossy  & 0.85  & $1.1 \times 10^6$ & 3.81 & 1.68  & 6.04  & 200 & 187.92 & 79.2 \\
Lossy Geometry  & 0.85 & $1.1 \times 10^6$ & 3.81 & - & 4.19 & 200 & 191.62 & 116.2 \\             
\hline
\end{tabular}
\end{table*}

\revise{As shown in Table \ref{table:table8}, lossless, lossy, and lossy geometry compression reduce transmission bandwidth by factors of 3, 21, and 31, respectively. For two-vehicle communication (${N_{car}} = 2$), the corresponding channel margin ratios are 67.5\%, 94.0\%, and 95.8\%. These results confirm that point cloud compression effectively alleviates bandwidth demands, with lossy geometry compression achieving the lowest bandwidth and highest margin. When scaled to larger communication scenarios (${N_{car}} = 11$), the channel margin ratios for lossy and lossy geometry compression remain at 39.6\% and 58.1\%, respectively, underscoring the advantage of discarding reflectance encoding in large-scale cooperative perception.}

\subsubsection{\revise{Evaluation for Compression Transmission on DAIR-V2X-V}}
\revise{In real V2X scenarios, we assessed the effectiveness of lossy geometry compression and transmission on the DAIR-V2X-V dataset, using metrics such as compression ratio, coordinate Bpp, and transmission bandwidth. The LiDAR sensor in this dataset operates with 40 beams at a 10 Hz capture frequency, producing approximately $7.2 \times 10^5$ points per second. As reported in Table \ref{table:table9}, PCC-S-C12, PCC-S-C11, and PCC-S-C10 achieve compression ratios of 24.55, 45.39, and 98.74, respectively, reducing transmission bandwidth by factors of 34, 94, and 373. These findings demonstrate that the PCC-S lossy geometry compression substantially reduces transmission costs in practical V2X environments, enabling real-time sharing of LiDAR data.}

\begin{table}[!t]
\centering
\caption{\revise{Quantization results of lossy geometry compression on the DAIR-V2X-V dataset. The key metrics include compression ratio, coordinate Bpp, and transmission bandwidth.}}
\label{table:table9}
\setlength{\tabcolsep}{3pt}
\begin{tabular}{c|c|c|c|c}
\hline
\multirow{2}{*}{Data Type} & \multirow{2}{*}{\shortstack{Points Per \\Second}} & \multirow{2}{*}{\shortstack{Compression \\Ratio}} & \multirow{2}{*}{\shortstack{Coordinate \\Bpp}} & \multirow{2}{*}{\shortstack{Bandwidth$\downarrow$ \vspace{-0.1em} \\ (Mbps)}}\\
  &  &  &  &   \\
\hline
\hline
Raw Data & $7.2 \times 10^5$ & - & - & 74.88  \\
PCC-S-C12 & $5.4 \times 10^5$ & 24.55 & 4.01 & 2.17 \\
PCC-S-C11 & $3.7 \times 10^5$ & 45.39 & 2.13 & 0.79 \\
PCC-S-C10 & $2.1 \times 10^5$ & 98.74 & 0.96 & 0.2 \\             
\hline
\end{tabular}
\end{table}

\subsection{Discussion of Limitation and Future Work}
For non-reflectance compressed data, we supplemented the missing reflectance information using the RP module and RKD, thereby enhancing detection robustness. Nevertheless, as shown in Table \ref{table:table1}, the APs for moderate-level cars and bicycles in KITTI still lag behind those achieved on raw data. A likely reason is that our RPKD framework emphasizes transferring detection knowledge via DKD without explicitly enhancing the geometry of lossy compressed point clouds.

\revise{Moreover, this study represents an initial step toward compression transmission-based collaborative perception, with the benchmark task limited to single-view 3D object detection. In real-world collaborative scenarios, connected vehicles will need to integrate locally collected raw data with decoded compressed data to expand perception range and improve detection accuracy. Therefore, effectively leveraging heterogeneous data of varying quality for registration, fusion, and 3D object detection remains a critical open challenge.}

In future work, we plan to develop geometric enhancement modules at the receiver to recover missing spatial details, further strengthening the robustness of 3D object detection on non-reflectance compressed data. \revise{Beyond this, we aim to advance collaborative 3D object detection by jointly exploiting raw and enhanced compressed data, thereby pushing forward the development of compression transmission-based collaborative perception.}

\section{Conclusion}
\label{sec:conclusion}
We have presented a 3D object detection framework with reflectance prediction-based knowledge distillation (RPKD) to improve detection accuracy for non-reflectance compressed point clouds. During training, the RCM module assigns the average reflectance of raw point-cloud voxels to corresponding compressed points based on their spatial relationships, providing essential prior labels for reflectance reconstruction. By treating non-reflectance compressed point clouds as weak data, the geometry-based RP module plays a critical role in accurate reflectance prediction and object detection within compressed point-cloud detectors. \revise{Building on CDTS, the raw point-cloud detector with the same structure transfers distillation knowledge through RKD and DKD, significantly boosting the robustness of compressed point-cloud detectors. Experimental results on the KITTI and DAIR-V2X-V datasets} demonstrate that our RPKD framework consistently improves detection performance across various code rates and backbone networks.

\bibliographystyle{IEEEtran}
\bibliography{bib}

\vspace{11pt}
\bf{}\vspace{-33pt}
\begin{IEEEbiography}[{\includegraphics[width=1in,height=1.25in,clip,keepaspectratio]{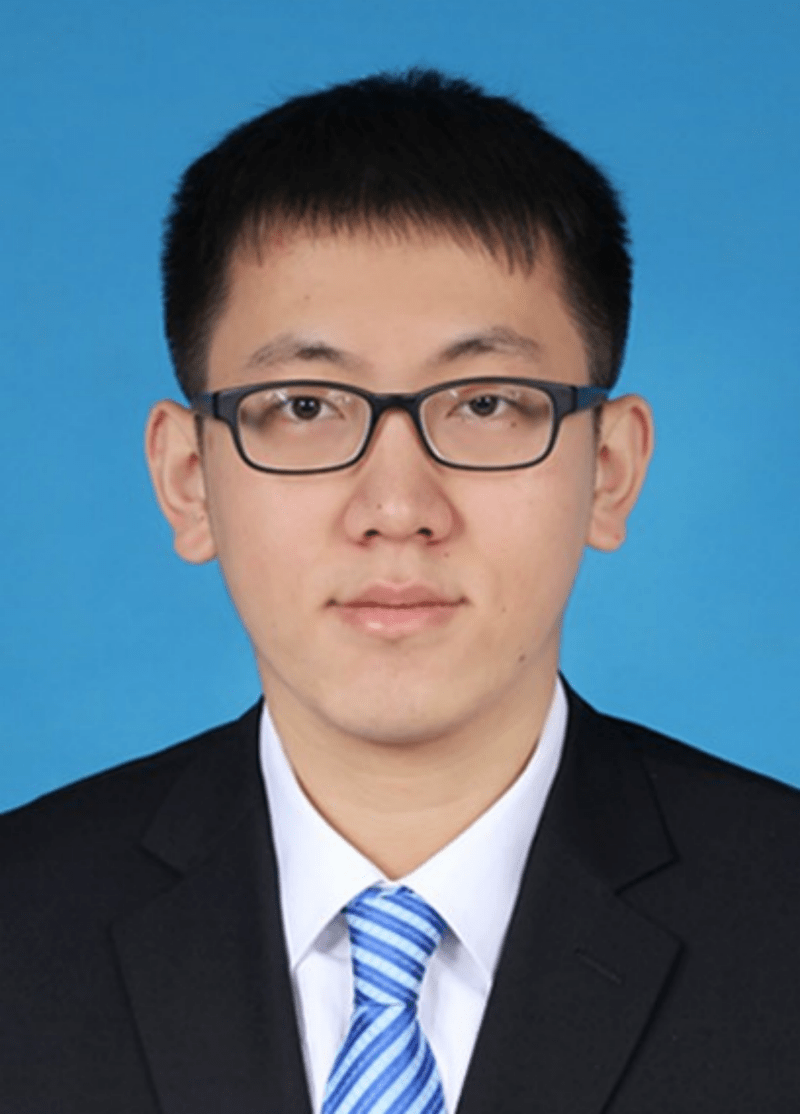}}]{Hao Jing}
received the B.E. and M.E. degrees from the University of Science and Technology Beijing, China, in 2014 and 2017, respectively. He is currently pursuing the Ph.D. degree with Taiyuan University of Science and Technology, China. His research interests include 3D object detection and computer vision.
\end{IEEEbiography}

\begin{IEEEbiography}[{\includegraphics[width=1in,height=1.25in,clip,keepaspectratio]{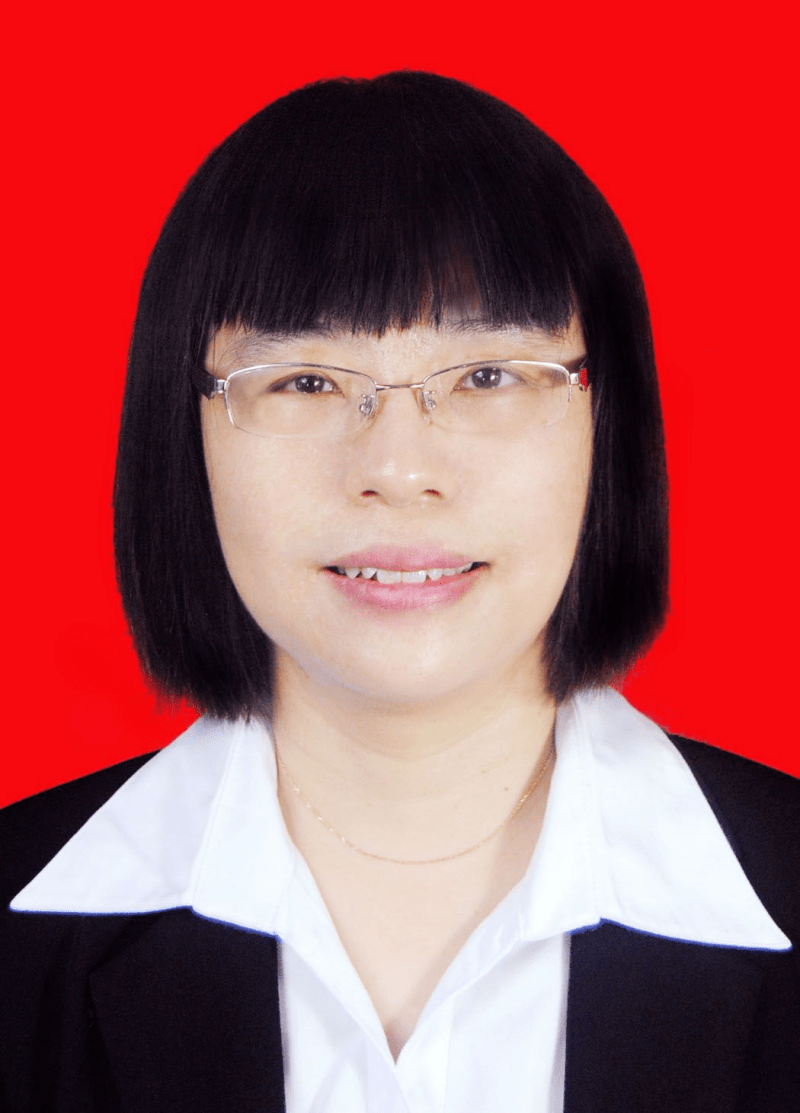}}]{Anhong Wang}
received the B.S. and M.S. degrees from Taiyuan University of Science and Technology, China, in 1994 and 2002, respectively, and the Ph.D. degree from the Institute of Information Science, Beijing Jiaotong University (BJTU), in 2009. She became an Associate Professor with TYUST in 2005 and became a Professor in 2009. She is currently the Director of the Institute of Digital Media and Communication, Taiyuan University of Science and Technology. Her research interests include image coding, video coding, and visual intelligence.
\end{IEEEbiography}

\begin{IEEEbiography}[{\includegraphics[width=1in,height=1.25in,clip,keepaspectratio]{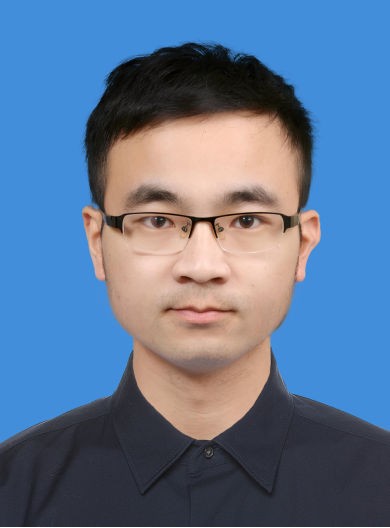}}]{Yifan Zhang}
received the B.E. degree from the Huazhong University of Science and Technology (HUST), the M.E. degree from Shanghai Jiao Tong University (SJTU), and the Ph.D. degree from the Department of Computer Science, City University of Hong Kong. Since 2025, he has been a lecturer at Shanghai University. His research interests include deep learning and 3D scene understanding.
\end{IEEEbiography}

\begin{IEEEbiography}[{\includegraphics[width=1in,height=1.25in,clip,keepaspectratio]{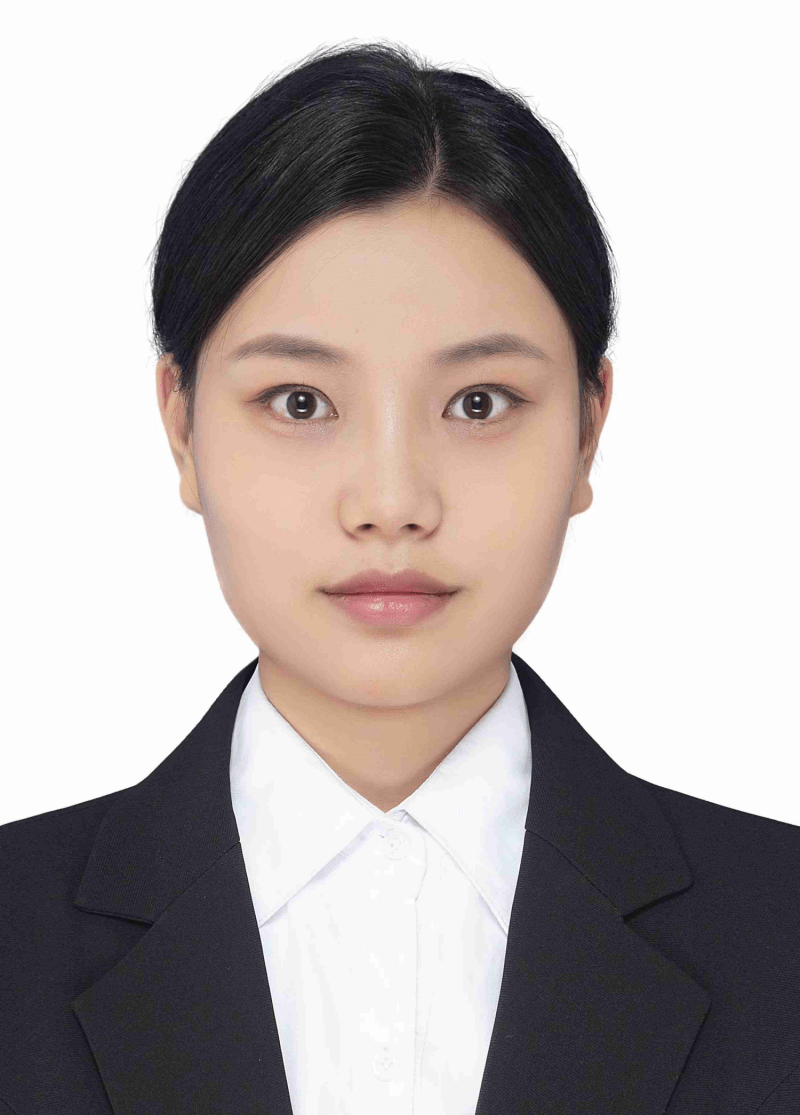}}]{Donghan Bu}
received the B.E. and M.E. degrees from Taiyuan University of Science and Technology, China, in 2017 and 2021, respectively, where she is currently pursuing the Ph.D. degree. Her research interests include 3D point cloud processing and computer vision.
\end{IEEEbiography}

\begin{IEEEbiography}[{\includegraphics[width=1in,height=1.25in,clip,keepaspectratio]{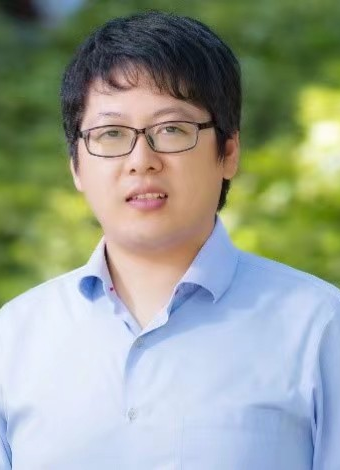}}]{Junhui Hou}
(Senior Member, IEEE) is an Associate Professor with the Department of Computer Science, City University of Hong Kong. He holds a B.Eng. degree in information engineering (Talented Students Program) from the South China University of Technology, Guangzhou, China, an M.Eng. degree in signal and information processing from Northwestern Polytechnical University, Xi’an, China, and a Ph.D. degree from the School of Electrical and Electronic Engineering, Nanyang Technological University, Singapore. His research interests are multi-dimensional visual computing.

Dr. Hou received the Early Career Award from the Hong Kong Research Grants Council in 2018 and the NSFC Excellent Young Scientists Fund in 2024. He has served or is serving as an Associate Editor for \textit{IEEE Transactions on Visualization and Computer Graphics}, \textit{IEEE Transactions on Image Processing}, \textit{IEEE Transactions on Multimedia}, and \textit{IEEE Transactions on Circuits and Systems for Video Technology}. 
\end{IEEEbiography}

\vfill
\end{document}